\documentclass{article} 
\usepackage{colm2024_conference}

\usepackage{graphicx}
\usepackage{enumitem}
\usepackage{wrapfig}
\usepackage{algorithm}
\usepackage{algpseudocode}
\usepackage{natbib}
\usepackage{makecell}
\usepackage{booktabs}
\usepackage{array}
\usepackage{amsmath}
\usepackage{amssymb}
\usepackage{amsfonts}
\usepackage{threeparttable}
\usepackage{multirow}
\usepackage{verbatim}
\usepackage{caption}
\usepackage{longtable}
\usepackage{ragged2e}      
\usepackage{array}         
\usepackage{supertabular}
\usepackage{CJKutf8}
\usepackage{array} 
\usepackage[utf8]{inputenc} 
\usepackage[T1]{fontenc}
\usepackage[french,vietnamese,mongolian,greek,english]{babel}
\usepackage{pifont}
\usepackage{xcolor}
\usepackage{enumitem}
\usepackage{tablefootnote}
\usepackage{xspace}
\usepackage{textcomp}
\usepackage{makecell}
\usepackage{lscape} 
\usepackage{siunitx}
\usepackage{listings}
\usepackage{xcolor}
\usepackage{colortbl}
\definecolor{kuaishoublue}{HTML}{6D9EEB}
\hypersetup{
    colorlinks=true,   
}
\definecolor{dt}{gray}{0.7}
\usepackage{pifont}       
\usepackage{bbding}       
\usepackage{fontawesome}
\newcolumntype{L}[1]{>{\raggedright\arraybackslash}m{#1}}

\usepackage{scrextend}

\usepackage{tgpagella}
\usepackage{latexsym}
\usepackage[T1]{fontenc}
\usepackage{microtype}
\definecolor{mydarkblue}{rgb}{0,0.08,0.45}
\definecolor{citecolor}{HTML}{0071BC}
\usepackage{url}            
\usepackage{nicefrac}       
\usepackage{changepage}
\usepackage{xargs}          
\usepackage{wrapfig,lipsum,booktabs}
\usepackage{longtable}
\usepackage{subcaption}
\usepackage{endnotes}
\usepackage{fancyvrb}
\usepackage{fvextra}
\usepackage{pgfplots}
\usetikzlibrary{pgfplots.groupplots}
\pgfplotsset{compat=1.3}
\usepackage{tikz}
\usetikzlibrary{patterns}

\usepackage[most]{tcolorbox}
\usepackage[capitalize,noabbrev]{cleveref}
\crefname{section}{Section}{\S\S}
\Crefname{section}{Section}{\S\S}
\crefname{table}{Table}{Tables}
\crefname{figure}{Figure}{Figures}
\crefname{algorithm}{Algorithm}{}
\crefname{equation}{eq.}{}
\crefname{appendix}{Appendix}{}
\crefformat{section}{Section #2#1#3}
\usepackage{multicol}
\usepackage{tcolorbox}

\usepackage{titlesec}
\titleformat*{\section}{\large\bfseries}
\definecolor{blue1}{HTML}{196ab1}
\definecolor{blue2}{HTML}{4886c1}
\definecolor{blue3}{HTML}{5e9bd6}
\definecolor{blue4}{HTML}{77b1e2}
\definecolor{blue5}{HTML}{bdd930}
\definecolor{blue6}{HTML}{dfebf6}

\definecolor{red1}{HTML}{de512c}
\definecolor{red2}{HTML}{f2642d}
\definecolor{red3}{HTML}{f68f58}
\definecolor{red4}{HTML}{febf92}
\definecolor{red5}{HTML}{f8e9c8}

\title{Kwai Keye-VL 1.5 Technical Report}

\author{
\bf Keye Team, Kuaishou Group}

\begin{document}

\maketitle

\begin{abstract}
In recent years, the development of Large Language Models (LLMs) has significantly advanced, extending their capabilities to multimodal tasks through Multimodal Large Language Models (MLLMs). However, video understanding remains a challenging area due to the dynamic and information-dense nature of videos. Existing models struggle with the trade-off between spatial resolution and temporal coverage when processing video content. We present Keye-VL-1.5, which addresses fundamental challenges in video comprehension through three key innovations. First, we introduce a novel Slow-Fast video encoding strategy that dynamically allocates computational resources based on inter-frame similarity, processing key frames with significant visual changes at higher resolution (Slow pathway) while handling relatively static frames with increased temporal coverage at lower resolution (Fast pathway). Second, we implement a progressive four-stage pre-training methodology that systematically extends the model's context length from 8K to 128K tokens, enabling processing of longer videos and more complex visual content. Third, we develop a comprehensive post-training pipeline focusing on reasoning enhancement and human preference alignment, incorporating a 5-step chain-of-thought data construction process, iterative GSPO-based reinforcement learning with progressive prompt hinting for difficult cases, and alignment training. Through extensive evaluation on public benchmarks and rigorous internal human assessment, Keye-VL-1.5 demonstrates significant improvements over existing models, particularly excelling in video understanding tasks while maintaining competitive performance on general multimodal benchmarks.
\end{abstract}

\begin{figure*}[ht]
\centering
\includegraphics[width= 1\linewidth]{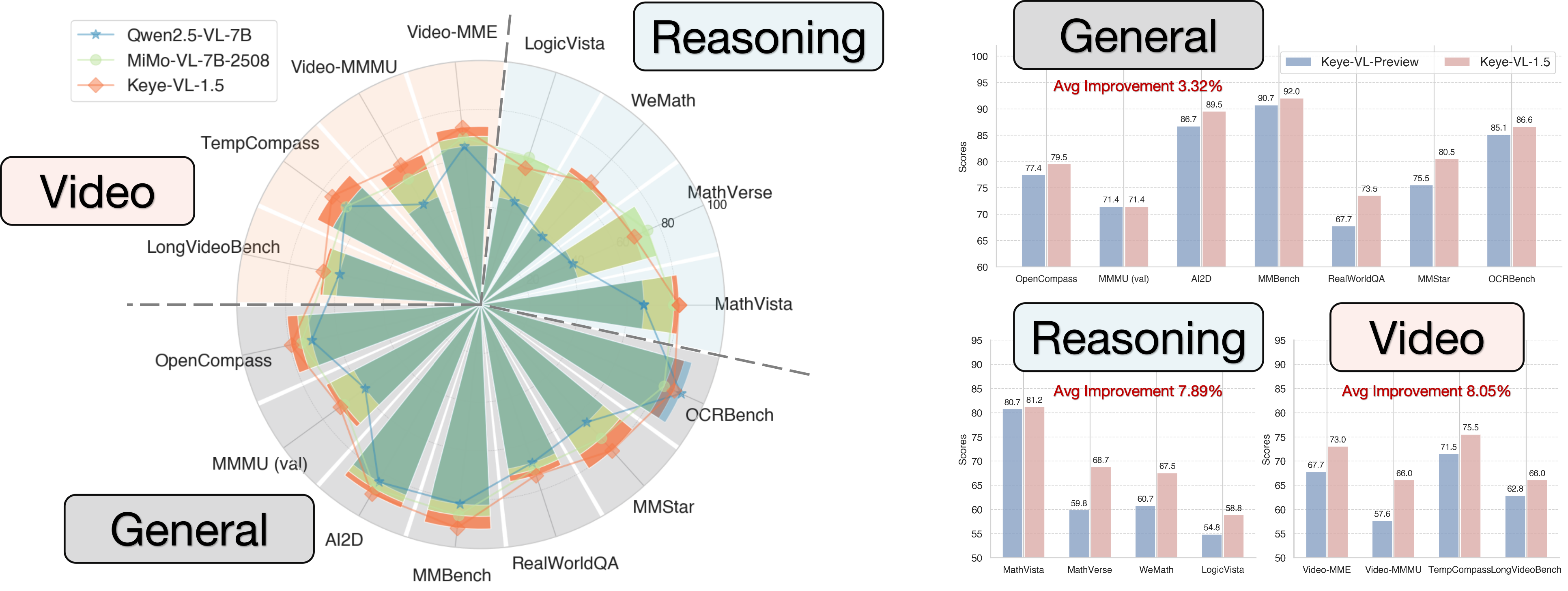}
\caption{\textbf{Benchmark Performance of Kwai Keye-VL-1.5.} Keye-VL-1.5-8B establishes new state-of-the-art performance among models of similar scale, demonstrating superior results on video-centric benchmarks while maintaining competitive performance on general multimodal and reasoning tasks. Compared to Keye-VL-Preview, this version shows significant improvements across all three evaluation dimensions, validating the effectiveness of our training approach.}
\end{figure*}

\newpage
{
  \setstretch{0.7}
  \tableofcontents
  \noindent\hrulefill
}
\newpage

\section{Introduction}

In recent years, Large Language Models (LLMs)(\cite{llama,phi3,ernie,emu3}) have experienced rapid development, ushering in a new era of artificial intelligence with their powerful capabilities in understanding (\cite{llama4,robo}), generation (\cite{qwen3,seed_1_5}), and linguistic reasoning (\cite{deepseek_r1, deepseek_v3}). This wave has also driven the rapid advancement of Multimodal Large Language Models (MLLMs) \cite{o3, mmict,expand,4o,kimi_vl,align_kd,vita_1_5,free,rain,cheap,seed_1_5_vl,team2025kwai,zhang2025thyme}), which extend powerful language capabilities to the visual domain, enabling the execution of complex tasks such as visual question answering (\cite{llava,internvl}), detailed image description (\cite{feast,eve,li2025denseworld1m}), object localization (\cite{qwen2_5_vl, mllm_selector}), and visual reasoning (\cite{o3, openthink, open_reasoner}).

Despite significant progress in static image understanding, video understanding remains a major challenge. Video content is inherently more dynamic and information-dense than static images, requiring models to process temporal relationships and sequential information while managing the fundamental trade-off between temporal coverage and spatial resolution. Existing approaches typically employ uniform frame sampling under fixed resolution constraints, which leads to suboptimal performance when fine-grained visual details and temporal consistency are required for content understanding (\cite{long_vita,video_llava,video_rag,gemini_rob,qwen2_5_vl}).

To address these limitations, we propose \textbf{Keye-VL-1.5}, an 8-billion parameter multimodal foundation model that achieves state-of-the-art performance in video understanding while maintaining robust capabilities in general vision-language tasks. Our contributions span three key areas: architectural innovations for efficient multimodal processing, progressive pre-training strategies, and comprehensive post-training methodologies.

\textbf{Architecture and Slow-Fast Video Encoding:} We propose a novel Slow-Fast video encoding strategy that dynamically allocates computational resources based on inter-frame similarity. Key frames with significant visual changes are processed through the Slow pathway at higher resolution, while relatively static frames are processed through the Fast pathway at lower resolution but with higher temporal coverage. This adaptive approach, guided by patch-based similarity functions, effectively addresses the trade-off between spatial detail and temporal breadth.

\textbf{Progressive Pre-training with Long Context Extension:} Our pre-training methodology comprises four carefully designed stages that progressively build multimodal capabilities. Beginning with cross-modal alignment and multi-task learning, we systematically extend the model's context length from 8K to 128K tokens during the annealing phase, enabling it to process longer videos and more complex visual content. This progressive approach ensures stable training while maximizing the utilization of the extended context window to enhance video understanding capabilities. The final model fusion stage combines models trained with different data mixtures to improve robustness and reduce bias.

\textbf{Post-training for Reasoning and Human Preference Alignment:} Our post-training process focuses on two critical aspects: enhancing reasoning capabilities and aligning with human preferences. We develop a comprehensive pipeline with three key components. First, we design a 5-step chain-of-thought reasoning data construction pipeline to generate high-quality cold-start data. Second, we employ the GSPO algorithm for verifiable reward-based reinforcement learning training. This includes progressive prompt sampling to handle difficult samples. Specifically, for samples where the model consistently fails during multiple rollouts, we provide varying levels of hints in the prompt to improve the efficiency of the rollouts. We use the RL model to generate better SFT data, and then perform the next round of RL training based on the SFT model, continuously iterating. Finally, we conduct alignment reinforcement learning training to enhance instruction following, response formatting, and preference alignment. This systematic approach ensures that Keye-VL-1.5 achieves excellent benchmark performance while providing responses that align with human expectations and preferences.

Through evaluation on public benchmarks and rigorous internal human assessment, we validate that Keye-VL-1.5 demonstrates significant improvements compared to existing models, particularly in video understanding tasks. Our work provides practical solutions for building next-generation multimodal models capable of complex video understanding and reasoning.

\begin{figure*}[t]
\centering
\includegraphics[width= 1\linewidth]{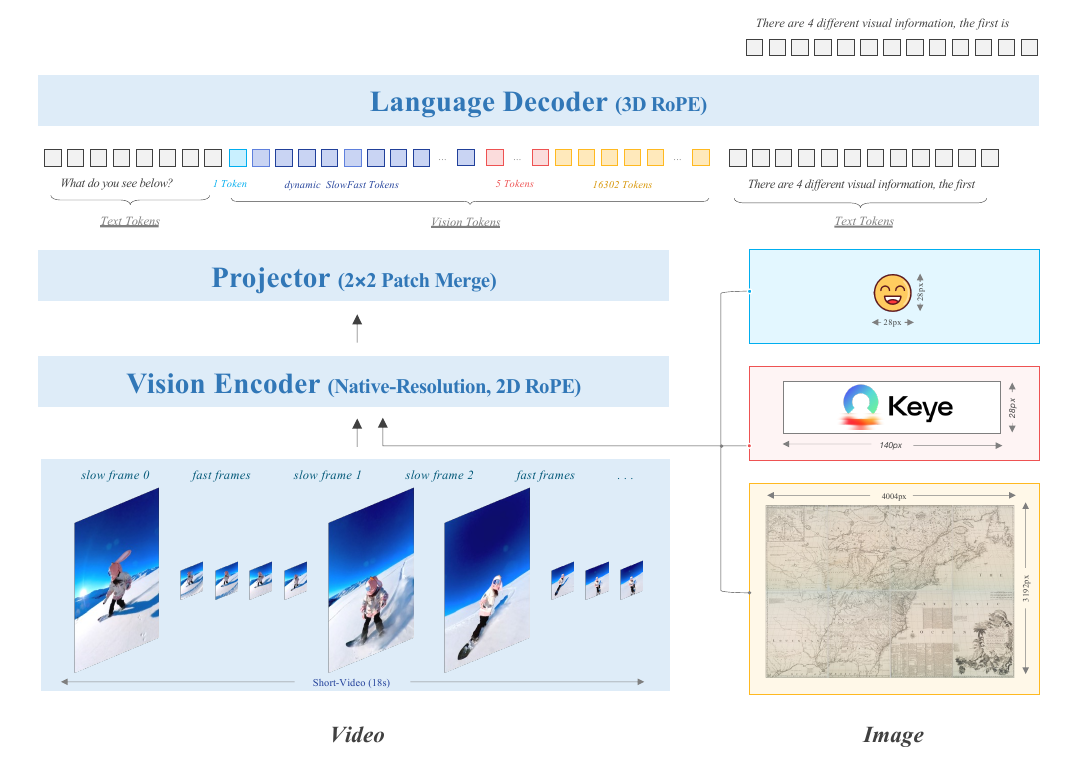}
\caption{\textbf{The Kwai Keye-VL-1.5 model architecture} is based on the Qwen3-8B language model and incorporates a vision encoder initialized from the open-source SigLIP. It supports SlowFast video encoding and native dynamic resolution, preserving the original aspect ratio of images by dividing each into a 14x14 patch sequence. A simple MLP layer then maps and merges the visual tokens. The model uses 3D RoPE for unified processing of text, image, and video information}
\label{fig:arc}
\end{figure*}

\section{Model Architecture}
Figure~\ref{fig:arc}  gives a high-level overview of our Keye-VL-1.5, which follows a classic MLLM architecture that includes three key components: a Vision Transformer (ViT), a MLP projector, and a language decoder.
For ViT component, we apply the open-source SigLIP-400M-384-14 \footnote{\url{https://huggingface.co/google/siglip-so400m-patch14-384}} as our vision encoder to extract vision information.
For LLM component, we employ the widely used Qwen3-8B as our language decoder, to provide the universal world semantic knowledge understanding capabilities.
For the projector, we randomly initialize its parameters and fully pre-training it at the Stage 1.
In the following sections, we provide our key upgrades, data pipeline and training recipes.

\subsection{Vision Encoder with Native-Resolution}
In past years, many MLLMs efforts have adopted the well-trained fixed-resolution ViTs as their vision encoders, such as ViT-bigG (\cite{vit_big}
), SigLIP-400M (\cite{siglip}) and others.
However, unlike pre-trained CLIP-based ViTs (\cite{clip}
) that only handle coarse-grained image-caption matching task during training, MLLMs often tackle various finer-grained generation tasks, existing a large gap between them. 
Therefore, we anticipate that our ViT will possess the following capabilities: during processing, images and videos maintain their structural integrity and all details are preserved.

To this end, there are some pioneer MLLMs exploring native-resolution ViT in recent years, such as Qwen2.5-VL, Seed-VL-1.5, Kimi-VL, etc.
In Keye-VL-1.5, we also implement a native-resolution ViT, to naturally process images at original resolution, avoiding some complex and redundant image splicing/splitting operations (e.g., MiniCPM2 (\cite{minicpm_v})).
Specifically, our ViT is initialized by the SigLIP-400M-384-14, a fixed-resolution variant with absolute learnable position embeddings to inject the spatial information.
According to it, we first employ interpolation techniques to extend fixed-length learnable position embeddings into resolution-adaptive position embeddings, enabling our basic native-resolution modeling while preserving the pretrained workflow.
Afterwards, to further enhance extrapolation capabilities for positional encoding along visual dimensions, we introduce 2D Rotary Position Embedding (RoPE) to strengthen the visual information modeling. 
In our trial experience, we observe that incorporating 2D RoPE significantly improves the model's performance on high-resolution image.
Finally, building upon the two types of position embeddings, we incorporate the NaViT packing with FlashAttention techniques to continue training our ViT across images with varying resolutions.

During the ViT pre-training procedure, we optimize our native-resolution modifications via SigLIP loss function (the text tower is also from SigLIP-400M-384-14). We use the same distribution data as the downstream MLLM for training, including a total of 500B Tokens from open source data DataComp (\cite{datacomp}), LAION (\cite{laion}), CC12M (\cite{cc12m}), PD12M (\cite{pd12m}), COCO (\cite{coco}) and other in-house data.

\begin{figure*}[t]
\centering
\includegraphics[width= 1\linewidth]{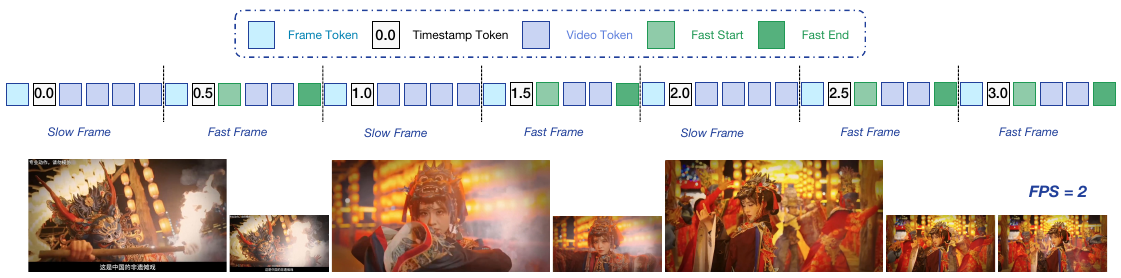}
\caption{\textbf{A SlowFast video encoding demonstration}: the Slow processes a smaller number of frames at higher resolution, while the Fast handles more frames at lower resolution.}
\label{fig:slowfast}
\end{figure*}

\subsection{Visual Encoding}
To guarantee that our language decoder can perceive enough visual signals to understand images and videos in detail, we devise different modeling strategies for them:
%
\begin{itemize}[leftmargin=*, noitemsep]
\renewcommand\labelitemi{$\diamond$}
\item \textbf{Native-Resolution Image Encoding:} for images encoding with different resolutions, we set the total number of tokens for each image to 20,480 (at LLM side), which can cover images with more than tens of million pixels and is sufficient to help the model to see the enough details of images.
\item \textbf{Slow-Fast Video Encoding:} for the video encoding with vary FPS, resolutions and duration, linearly increasing any of these factors would lead to a sharp increase in the token budget on the LLM side, thus making it challenging to strike a balance between performance and cost.
%
To our knowledge, most existing MLLMs typically adopt a fixed number of frames and accordingly reduce the resolution of each frame to meet token budget limitations. 
%
Following the paradigm, Qwen-2.5-VL further proposes 2D convolution technique to merge the adjacent frames, aiming to enable the LLM decoder to perceive more video signals within a fixed frame count.
%
Nevertheless, under the uniform frame sampling strategy, although many adjacent frames may be highly similar, there can still be some cases where consecutive frames show significant differences, especially when sampling-interval is larger, a person is moving  or viewpoint is shifting.
As a result, the rough 2D convolution merging technique maybe unfriendly to effective video understanding, since it relies on overly strong assumptions.
Considering the inherent characteristics of video: where adjacent frames are mostly similar yet sometimes significant changes, we propose a SlowFast video encoding strategy:
\begin{itemize}[leftmargin=*, noitemsep]
\renewcommand\labelitemi{$\diamond$}
\item \textbf{Slow Pathway}: This pathway is designed to capture visual information from rapidly changing frames. It operates at a lower number of frames but with higher resolution.
\item \textbf{Fast Pathway}: In contrast, the Fast Pathway captures subtle changes visual signal from relatively static frames. It uses a higher number of frames but at a lower resolution.
\end{itemize}
To identify the slow/fast frames from the video, we first devise a patch-based similarity function to extract them: (1) The first frame is always defined as a slow frame; (2) For each subsequent frame, if its patch similarity with the latest slow frame exceeds 95\%, it is marked as a fast frame; otherwise, it is marked as a new slow frame.
%
After obtaining the slow and fast frames, we set the fast frame's token budget to 30\% of a slow frame's budget to balance the trade-off between frame numbers and the total token budget. Then, we utilize a binary search technique to precisely calculate the number of tokens per slow frame under the total token budget limitation (e.g., 75,000 tokens in Keye-VL-1.5).
Meanwhile, to more clearly identify the boundaries and timestamp information between Slow and Fast frames, we introduce additional special tokens along with absolute timestamps to guide the model during learning, as shown in Figure~\ref{fig:slowfast}.

%



%
%

\end{itemize}
%


\section{Pre-Training}
In this section, we first describe the construction of the pre-training dataset, followed by an overview of the overall training pipeline and configuration.

\subsection{Data Pipeline}
In our data construction pipeline, we have assembled a diverse, high-quality corpus with exceeding 1 trillion tokens to support our models training, sourced from both public datasets and proprietary in-house data.
Generally, our training data encompasses six primary categories: Image Caption, OCR \& VQA, Grounding \& Counting, Interleaved, Video Understanding and Pure Text data.
To ensure these overall data quality, we have designed customized filtering mechanisms tailored to the characteristics of each data category.
For large volumes of medium-quality data, we employ CLIP (\cite{clip}) scores for preliminary filtering. 
For smaller amounts of high-quality data, we utilize open-source MLLMs as discriminators for data selection.
Additionally, we also conduct rigorous image-based deduplication operation, to avoid the potential data leakage between our training corpus and evaluation benchmarks (\cite{corruption}).
Specifically, we identify highly similar images, then remove these near-duplicates from the dataset.
%
%
In the following sections, we provide detailed descriptions of each category of our data.

\subsubsection{Image Caption Data}
Image caption task provides the fundamental world knowledge to establish a mapping relationship between visual features and linguistic concepts by pairing image with textual descriptions.
Based on large-scale caption data, our model gains the ability to perceive and comprehend a broad, rich spectrum of world knowledge, such as real-world physical principles and cultural conventions.
Although we can public access many diverse Chinese and English open-source caption data source, such as LAION (\cite{laion}), DataComp (\cite{datacomp}) and Coyo (\cite{coyo}),
the quality of such data is often unreliable, as it typically only undergoes simple crawler-based matching.

To alleviate such data noise, we conduct strict similarity-based filtering pipeline to control the data quality, e.g., scoring the raw rigorous image-caption pair by a CLIP model.
In practice, to ensure data quality, we retain high-similarity image-caption pairs (e.g., CLIP score > 0.9) while leveraging filtered low-quality open-source image data and our in-house image data through a re-captioning pipeline.
During the re-caption, we utilize several MLLMs (Qwen2.5-VL 72B (\cite{qwen2_5_vl}), Tarsier2 (\cite{tarsier2}), GPT-4o (\cite{4o}), Gemini1.5-pro (\cite{gemini}) and others) to generate the synthesis caption for vary resolution images and image category information.
In our experience, we find that recaption data generated by different MLLMs can be very helpful for fine-grained image understanding.

Further, to avoid our model degenerate into a caption generators and hurt its instruction-following and complex reasoning abilities.
We implemented a data augmentation strategy with multiple-caption/question-answering pair to maintain our model’s general conversation and instruction capabilities:
\begin{itemize}[leftmargin=*, noitemsep]
\renewcommand\labelitemi{$\diamond$}
\item <image, caption, [eos], question, answer> format data: training our model to seamlessly transition from generating captions to accurately answering follow-up questions, thereby strengthening contextual understanding and task continuity.
\item <image, question, answer, [eos], caption> format data: Reverses the task order, requiring the model to answer before describing, which helps break the tendency to default to caption generation and improves task-switching flexibility and instruction sensitivity.
\item Instruction-following image captioning/QA: We first provide dozens of images as input, then randomly ask questions or generate captions corresponding to specific images.
\end{itemize}
Besides, to improve our model robustness and faithfulness, we proactively inject some `trap questions' that refer to non-existent or contradictory questions. These counterfactual data would encourage the model to ground its responses more accurately in visual content rather than textual priors.

\subsubsection{OCR \& VQA Data}
Optical Character Recognition (OCR) and Visual Question Answering (VQA) are vital tasks to encourage our model to distinguish the details of images.
By integrating OCR capabilities, the model can accurately extract and interpret textual information within images, while VQA task enables our model to comprehend and reason about visual content in a context-aware manner.
In order to build our capabilities in OCR and VQA, we have collected a large number of open-source data, such as Latex-Formula, hand-write text, real-world street views, charts, rich-text documents, multi-image OCR and so on.
Since most of the open-source datasets are in English, to further enhance the model's capability in Chinese OCR \& VQA tasks, we introduce multiple techniques for synthesizing in-house Chinese data:
\begin{itemize}[leftmargin=*, noitemsep]
\renewcommand\labelitemi{$\diamond$}
    \item \textbf{Synthesis with SOTA MLLMs:} To enhance the model's OCR capabilities, we extract images from both open-source and in-house image-text datasets to build our image repository, utilizing the text-dense images from it to synthesize comprehensive OCR dataset which covering diverse scenario. For VQA task, we first design a set of seed-questions and expand the initial question pool through self-evolution methods. Next, both images and their corresponding captions are fed into SOTA MLLMs to generate high-quality and diverse VQA data, such as utilizing Qwen2.5-VL-72B to generate multi-turn challenging question-answering pairs.
    \item \textbf{Rendering with Font Tools:} Considering the scarcity of high-quality open-source Chinese OCR data, we further leverage font rendering tools to synthesize high-quality OCR samples (includes (1) diverse image backgrounds/layout,  (2) semantic/non-semantic text, (3) multiple fonts styles/sizes and (4) vary image resolutions), which significantly enhances the model’s robustness for Chinese OCR recognition.
    \item \textbf{Structured Document and Code Understanding:} Further, we also perform complex text recognition tasks by using a vast codebase (e.g., Markdown, HTML, and other programming languages). By rendering codes/documents that preserve their original layout, we could create elaborate OCR tasks, such as reconstructing source code from an image or completing missing code at specific locations, thereby training the model to understand textual hierarchy and structure.
    \item \textbf{Instruction Following OCR:} Moreover, to enhance our model's capability to follow specific OCR instructions (e.g., ``extract only the text from the third column''), we built an instruction-following OCR dataset. Each sample consists of a character-matrix image paired with a Chinese instruction, covering thirteen classes of reading and locating templates (e.g., row/column extraction in four directions and their combinations). This dataset is enriched with diverse text sources, noise injection (English, Japanese, Korean, symbols, numbers, uncommon characters, and emojis).
\end{itemize}

\subsubsection{Grounding \& Counting Data}
Object grounding is one of the fundamental abilities of MLLMs(~\cite{qwen2_5_vl,seed_1_5}), which enables our model to establish a direct connection between temporal/visual information and text semantics, as shown in the Table~\ref{Grounding}.
In Keye-VL-1.5 objective grounding, we primarily utilize three object localization forms: center points, bounding boxes, and polygons.
Their coordinates are strictly typed as integers and normalized to the range [0, 1000) for different resolution images, .
In general, we mainly employ the RefCoCo (\cite{refcoco}), VisualGenome (\cite{visual_genome}), TolokaVQA (\cite{toloka_vqa}) as our grounding data source, and the PixMo (\cite{pixmo}) as our counting data source.
For the in-house grounding data generation, we use other MLLMs (e.g., Gemini 2.5 Pro, Qwen-2.5-72B) to extract the answer area bounding boxes of corresponding document questions.
To filter the incorrect, missing, or ambiguous annotation grounding data, we utilize the CLIP and Qwen-2.5-7B to select the higher-score points/boxes/polygons as our training data, i.e., extracting the corresponding grounding area from the image to compute its similarity with the target objective text.

For temporal grounding data, we construct a three-step coarse-to-fine-grained data synthesis pipeline based on our massive short-videos base.
In the first step, we employ the TEMPURA~\cite{cheng2025tempura} to process a given short-video as several event video clips with their temporal captions.
Next, to alleviate the ``repetitive collapse'' in redundant or meaningless descriptions issue of raw TEMPURA outputs, we apply the SOTA MLLMs as a filter to identify and remove such low-quality, repetitive event video clips to obtain reliable temporal grounding captions.
At last, according to those captions, we further utilize the Gemini 2.5 Pro to enrich our database to generate a series of logical question-answering pairs about timestamps, which could empower our model's understanding of temporal causality relationships.
In this way, our pipeline ensures our model not only describes what happens in a video, but also understands and reasons about when and why.

\begin{table}[t!]
\footnotesize
\centering
\setlength{\tabcolsep}{0pt}{
\resizebox{0.99\textwidth}{!}{
\begin{tabular}{lc}
\toprule
\multicolumn{2}{c}{\textbf{objective center points}} \\
\midrule
Example& <|point\_start|>[[x1, y1]]<|point\_end|> \\
Description& The [x1, y1] is the center point of queried objective.\\
\midrule
Example& <|point\_start|>[[x1, y1], [x2, y2]]<|point\_end|> \\
Description& Supporting multiple points for a single queried objective.\\
\midrule
Example& <|object\_ref\_start|>obj<|object\_ref\_end|><|point\_start|>[[x1, y1]]<|point\_end|> \\
Description& The [x1, y1] is the center point of `obj'. \\
\midrule
\multicolumn{2}{c}{\textbf{objective bounding boxes}} \\
\midrule
Example& <|box\_start|>[[x1, y1, x2, y2]]<|box\_end|> \\
Description& The coordinates [x1, y1]/[x2, y2] denote the top-left and bottom-right point of box of queried objective.\\
\midrule
Example& <|box\_start|>[[x1, y1, x2, y2], [x3, y3, x4, y4]]<|box\_end|> \\
Description& Supporting multiple boxes for a single queried objective.\\
\midrule
Example& <|object\_ref\_start|>obj<|object\_ref\_end|><|box\_start|>[[x1, y1, x2, y2]]<|box\_end|>\\
Description& Detecting the `obj' and its corresponding box.\\
\midrule
Example& <|ocr\_text\_start|>text<|ocr\_text\_end|><|box\_start|>[[x1, y1, x2, y2]]<|box\_end|> \\
Description& Identify the OCR results and its corresponding box.\\
\midrule
\multicolumn{2}{c}{\textbf{objective polygons}} \\
\midrule
Example& <|object\_ref\_start|>obj<|object\_ref\_end|><|polygon\_start|>[[[x1, y1], [x2, y2], [x3, y3]]]<|polygon\_end|> \\
Description& The coordinates [x1, y1], [x2, y2], ... represent polygon vertices of `obj', which arranged in clockwise order.\\
\midrule
Example& <|ocr\_text\_start|>text<|ocr\_text\_end|><|polygon\_start|>[[[x1, y1], [x2, y2], [x3, y3]]]<|polygon\_end|> \\
Description& Supporting the OCR results.\\
\midrule
\multicolumn{2}{c}{\textbf{temporal caption}} \\
\midrule
Example& <|clip\_time\_start|>[t1, t2]<|clip\_time\_end|> event-caption.\\
Description& [t1, t2] represents time grounding duration of the corresponding event.\\
\bottomrule
\end{tabular}
}
}
\caption{Grounding Label Assembling of Keye-VL-1.5.}
\label{Grounding}
\end{table}

\subsubsection{Interleaved Text-Image Data}
Instead of the learning task surrounding the single images, we also introduce a large amount of interleaved data to enhance our language decoder's longer multi-modal context modeling ability and longer sequence adaptation, e.g., 128K context modeling.
Actually, beyond modeling multi-image correlations, the interleaved data could contribute several critical advantages in pre-training: (1) Preservation of General Knowledge: It contains a wealth of universal knowledge, ensuring that the LLM module's core capabilities are not degraded during training, (2) Enhanced Vision-Language Alignment: By leveraging in-context learning, it helps the model better align visual and semantic signals in language model side, (3) Improved Generalization: The diverse and interleaved nature of the data strengthens the model's ability to reason across modalities and generalize to unseen tasks.
Besides the open-source interleaved data, we also build a large-scale in-house interleaved data generation pipeline.
Specifically, we focus on the two type of raw rich-text documents processing, the academic PDF data and structured knowledge data, especially the Science, Technology, Engineering, and Mathematics (STEM) data.
We collect a substantial amount of academic and knowledge-based PDF/structured data to render the text content into plain text format and insert the corresponding images at their original positions within the text.
In such a process, we conduct rigorous data protection strategies to ensure high-quality outputs. Our pipeline includes: (1) Garbled character recognition: identifying and removing garbled characters, (2) Low-resolution/broken image filtering: ensuring image quality, (3) Text-image similarity validation: ensuring semantic alignment between interleaved image-text.

\subsubsection{Video Data}
As a short-video and live-streaming service provider, the video understanding ability is the most important point of Kwai, such as understanding the video details, generating summaries, and expressing interesting implications.
To reach the goal, our video data are collected from multiple sources, including diverse open-source datasets (ShareGPT4V, Pandas and others) and a large-scale high-quality in-house video data.
Based on these videos, we conduct the following key pipelines to guarantee our data quality:
\begin{itemize}[noitemsep]
\renewcommand\labelitemi{$\diamond$}
    \item Interleaved video-ASR: For audio signals, we currently use speech-to-text tools (e.g., Qwen2.5-Omni (\cite{qwen2_5_omni})) to recognize them, and then form a interleaved style to connect images and audio to our model.
    \item Video recaption: With (optional) ASR results, we next utilize diverse public MLLMs to generate its caption under different FPS setting, such as 0.5/1/2.
    \item Frame-level OCR annotation: In order to ensure that our model does not miss any details in each frame, we further added a frame-level OCR task.
\end{itemize}
In addition to OCR and video captioning/QA tasks, we have designed a series of reasoning-enhanced tasks to help the model better understand contextual relationships in short videos. These include:
\begin{itemize}[leftmargin=*, noitemsep]
\renewcommand\labelitemi{$\diamond$}
\item Frame-level re-ordering: Given a set of shuffled video frames, our model is required to predict their original chronological order, which enhances its ability to grasp temporal progression and logical flow.
\item Multiple video matching: Provided with a group of related videos and a set of candidate videos, our model is required to identify the most contextually relevant candidate, which refines its understanding of semantic connections across different videos.
\end{itemize}

\begin{figure*}[t]
\centering
\includegraphics[width= 1\linewidth]{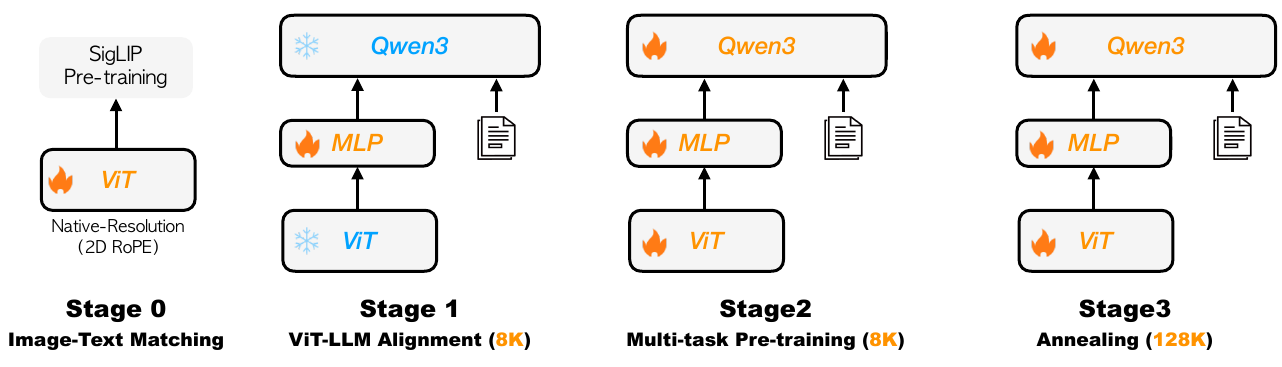}
    \caption{\textbf{The Kwai Keye-VL-1.5 pre-training pipeline,} featuring a four-stage progressive strategy: Image-Text Matching, ViT-LLM Alignment, Multi-task Pre-training, and Annealing with model merging.}
\label{fig:pretraining_stages}
\end{figure*}

\subsection{Training Recipe}
We employ a four-stage progressive training strategy to build a powerful multi-modal foundation model with strong vision-language alignment capabilities. The training pipeline, illustrated in Figure \ref{fig:pretraining_stages}, is meticulously designed to ensure that each stage has a clear and interconnected objective.

The Vision Transformer (\cite{vit}) (ViT) is initialized with weights from the \textit{siglip-so400m-patch14-384} model and undergoes continuous pre-training using the SigLIP (\cite{siglip}) contrastive loss function. This stage focuses on adapting the vision encoder to our internal data distribution. We incorporate native dynamic resolution processing (akin to NaViT (\cite{navit})), which preserves the original aspect ratio of images to the greatest extent possible. Additionally, 2D Rotary Position Embeddings (\cite{rope}) (RoPE) are integrated to enhance the model's extrapolation capabilities when processing images of varying resolutions.

\paragraph*{Stage 1: cross-modal alignment:}
The language model is initialized from Qwen3-8B (\cite{qwen3}). During this stage, the parameters of both the vision and language models are frozen. Training is focused on optimizing the projection MLP layer. With large-scale datasets, we establish a robust alignment between cross-modal features, laying the groundwork for the subsequent learning phase.

\paragraph*{Stage 2: multi-task pre-training:}
All model parameters are unfrozen for end-to-end optimization using a diverse set of multi-task training data. The data in this stage encompasses a wide range of common vision-language tasks, including Image Captioning, Optical Character Recognition (OCR), Grounding, Visual Question Answering (VQA), and interleaved image-text data. This process significantly enhances the model's fundamental visual understanding capabilities.

\paragraph*{Stage 3: annealing:}
This stage involves an annealing phase where the model is fine-tuned on a curated set of high-quality data. The primary goal is to address the issue of insufficient exposure to high-quality samples during the large-scale, broader training of Stage 2. Through optimized learning strategies and data mixtures, we further refine the model's nuanced understanding and capabilities.

\paragraph*{Sequence Length Extension to 128K:}
In Stage 1 and Stage 2, we limit the sequence length of each sample to 8,192 (8K), where Data Parallelism is adopted to effectively create large batch sizes. Zero-2 optimization strategy is applied to reduce memory overhead.
In the final annealing stage, we extend the context length of the model from 8,192 (8K) to 131,072 (128K). The RoPE inverse frequency of LLM side is reset from 1,000,000 to 8,000,000. The training data is concurrently enriched with high-quality long-context modalities, including long videos, long texts, and large-scale images. Additionally, we switch optimization strategy to Zero-1 and adopt Context Parallelism and Pipeline Parallelism to support long-context training. Under the 128K context length, our controlled experiments show that allocating 24\% of tokens to videos, 50\% to images, and the remaining 26\% to text strikes a good balance between visual capabilities (image and video understanding) and text capabilities.


\section{Post-Training}
\begin{figure*}[t]
\centering
\includegraphics[width= 1\linewidth]{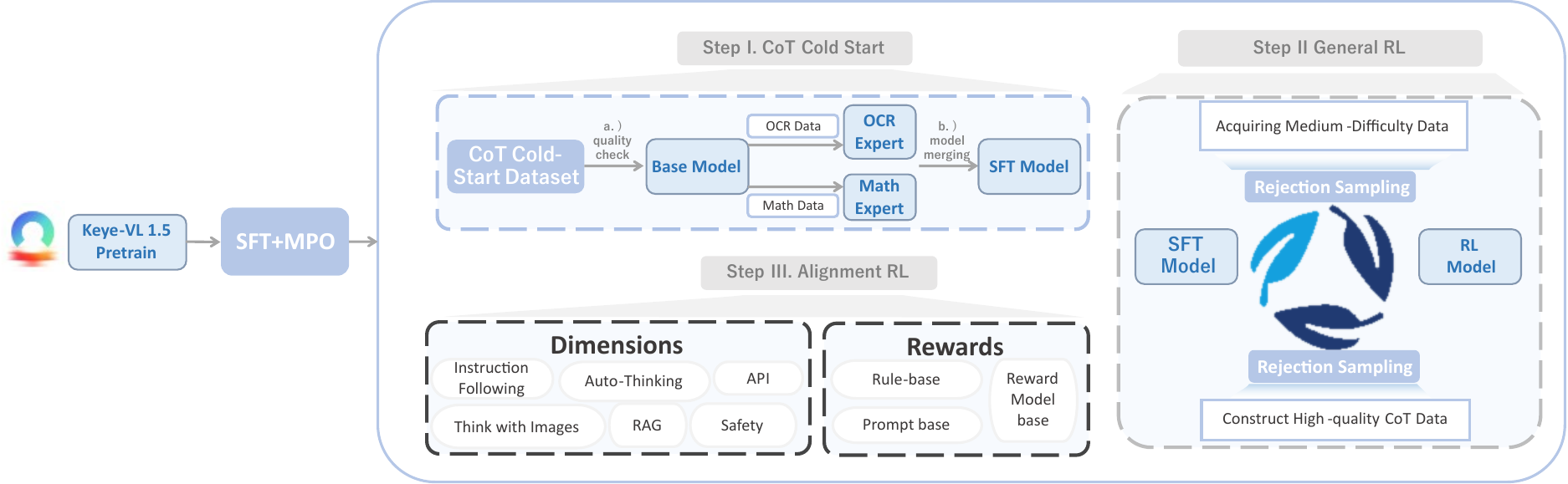}
\caption{\textbf{Post-Training Pipeline:} The post-training process includes non-reasoning stage and reasoning stage. The non-reasoning stage is composed of SFT and MPO training. The reasoning stage consists of three key steps: CoT Cold Start (we construct a five-step construction pipeline to generate high-quality CoT Cold-Start Dataset and apply model merging to refine model performance), General RL (we concentrate on improving Keye-VL-1.5's reasoning ability, applying GSPO, we propose progressive hint sampling to fully take advantage of hard problems and iteratively improve the cold-start and general RL model), and Alignment RL (improving Keye-VL-1.5's instruction following, format adherence, preference alignment and RAG ability with our reward system, we construct instruction following data, reasoning data and RAG data for RL training in this step).}
\label{fig:reasoning-training}
\end{figure*}

\subsection{Non-Reasoning Stage: SFT + MPO}

The SFT data candidate pool contains over 7.5 million multimodal QA samples. We employ the following construction methods to balance comprehensiveness and data quality. 

\begin{itemize}[leftmargin=*, noitemsep]
\renewcommand\labelitemi{$\diamond$} 
\item \textbf{To ensure task diversity,} we utilize the proprietary TaskGalaxy (\cite{taskgalaxy}) framework, which categorizes data across a comprehensive system of 70,000 distinct multimodal task types. We further construct a large amount of data for image/video grounding, counting, GUI, and multi-turn dialogue.
\item \textbf{To ensure the data's challenge,} MLLMs are employed to generate multiple reasoning paths for each data point. The complexity of each sample is then measured based on the correctness and length of these responses, allowing for the filtration of overly simple data. We increase the proportion of mathematical, logical reasoning, complex tasks, and long-context data.
\item \textbf{To ensure data reliability,} human annotators have meticulously crafted captions for the images and videos within the training set.
\end{itemize}
The training strategy involves a dynamic learning rate. In the later phases of training, the model undergoes an annealing process at a lower learning rate. Evaluations show this annealing step contributes approximately a 1\% performance improvement across both open-source and internal benchmarks.

Following SFT, the model undergoes MPO to continuously refine its performance.
The MPO dataset includes 250k open-source samples~\cite{mpo}, 150k text-only samples, and 26k human-annotated samples~\cite{mm_rlhf}. We perform multiple samplings using Keye-VL-1.5 on the above dataset, and construct multiple pairs of high-quality and low-quality samples using the reward model scores and human annotations.
The training strategy for this stage applies the MPO algorithm, utilizing the constructed paired preference data to optimize Keye-VL-1.5's overall performance.

\subsection{Keye-Reward Model}
Recognizing the importance of reward modeling for data quality evaluation and model training, we train our reward model based on the Keye-VL-preview for data filtering and reinforcement learning training. We adapt the Keye-VL-preview model to the reward modeling task with the SFT+RL training process.

\textbf{Data format}: The model input consists of the query, response A, and response B, along with the task definition guiding the model in evaluating the quality of response A and response B. Similar to Keye-VL-1.5's mix reasoning mode, our training data is composed of two formats: think and no\_think. For the no\_think mode, the model directly outputs the final judgment based on the input information. In the think mode, the model needs to evaluate the quality of response A and response B separately according to the predefined nine dimensions (such as Credibility, Correctness, Redundancy, Relevance, etc.), then generate a comprehensive evaluation. The mix reasoning mode enables our reward model to reason in terms of efficiency, accuracy, and interpretability.

\textbf{SFT recipe}: The SFT data includes open sourced preference datasets R1-Reward~(\cite{zhang2025r1}), MMPR~(\cite{mpo}), and manually labeled Keye-VL-preview sampling results. After SFT, we apply data where good responses are shorter than bad responses for annealing to overcome the reward model's preference for longer responses.

\textbf{RL recipe}: The RL data includes preference data consisting of wrong cases from Keye-VL-preview in the SFT dataset and right cases generated by larger MLLMs, as well as data from MMPR. In this stage, we carefully filter out data with excessively large length differences between positive and negative samples, using format reward and outcome reward as training signals.

We take our reward model to evaluate the quality of Keye-VL's sampling results, which are applied to update the training data and provide reward signals.
\subsection{LongCoT Cold-Start}
After large-scale SFT and MPO, we construct high-quality Long Chain-of-Thought (LongCoT) data for cold-start reasoning training, aiming to enhance Keye-VL's long CoT reasoning ability, serving as the starting point for subsequent reinforcement learning.

\subsubsection{Data Construction Pipeline}
\begin{figure*}[t]
\centering
\includegraphics[width= 1\linewidth]{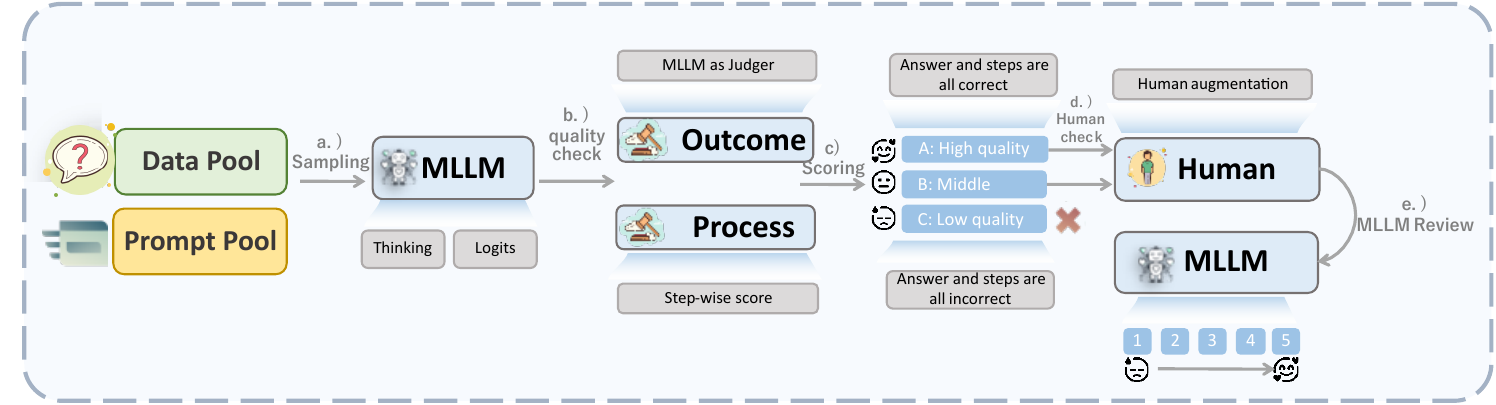}
\caption{\textbf{Overview of our five-step automated LongCoT data generation pipeline.} The pipeline begins with (a) sampling from data and prompt pools using MLLMs to generate thinking processes and logit information, followed by (b) quality assessment using MLLM as judge to evaluate both outcomes and reasoning processes with step-wise scoring, (c) categorization into three quality tiers (A: high quality, B: middle quality requiring human review, C: low quality to be discarded), (d) human augmentation for Category B samples and suspected redundant Category A samples, and (e) final MLLM review with dynamic quality scoring (1-5 scale) to determine optimal data utilization strategies. This comprehensive approach ensures both scalability and quality control in generating training data.}
\label{fig:pipeline}\end{figure*}

To address the challenge of acquiring high-quality training data for cold-start, we propose a comprehensive five-step automated pipeline for generating LongCoT data, as illustrated in Figure~\ref{fig:pipeline}. Our approach strategically leverages existing MLLMs to create diverse, high-quality reasoning chains while maintaining both scalability and cost-effectiveness. The pipeline systematically integrates automated generation, rigorous quality assessment, targeted human enhancement, and adaptive data utilization to ensure optimal training data quality across diverse domains and reasoning complexity levels.

\textbf{Multi-Source Data Collection and Enhancement:} Our data generation process begins with the systematic collection of multimodal QA data spanning multiple challenging domains. These domains include mathematical reasoning problems, STEM, OCR and document understanding tasks, visual grounding and object localization, counting, GUI scenarios, and domain-specific business applications. This comprehensive coverage ensures that our generated dataset captures the full spectrum of multimodal reasoning capabilities required for practical applications.

To enhance the complexity and diversity of the collected data, we employ proprietary MLLMs to perform sophisticated question rewriting and task merging operations. The rewriting process transforms simple, straightforward questions into more challenging variants that require deeper reasoning and multi-step problem solving. Additionally, we systematically combine related sub-tasks into comprehensive multi-task instructions, creating scenarios where models must demonstrate proficiency across multiple capabilities simultaneously. This enhancement strategy significantly increases the pedagogical value of each training sample while maintaining natural question flow and coherence.

\textbf{Multi-Path Reasoning Generation with Confidence Quantification:} For each enhanced QA pair, we generate multiple reasoning trajectories leveraging existing MLLMs. A pivotal component of our generation pipeline is the systematic extraction and quantification of model confidence at both the step-wise and holistic response levels. We compute granular confidence scores that capture the model's certainty in individual reasoning steps as well as the final answer. This confidence metadata serves as a crucial signal for downstream quality assessment and sample prioritization workflows, enabling us to systematically identify the most reliable and coherent reasoning chains from the generated candidate pool. Throughout the multi-round sampling process, we strategically select samples that exhibit diverse logical pathways while maintaining correctness, thereby enriching the diversity of reasoning patterns. Simultaneously, we implement a confidence-prioritized selection strategy, systematically favoring reasoning chains with higher logit-based confidence scores to optimize training sample quality.

\textbf{Comprehensive Two-Level Quality Assessment:} We implement a rigorous two-level quality assessment framework using proprietary MLLMs. This dual assessment strategy operates simultaneously on both answer correctness and reasoning process validity. At the answer level, our assessment framework incorporates flexible matching patterns specifically tailored to different task types and domains. The system supports sophisticated fuzzy matching capabilities and equivalent expression recognition, accommodating variations in phrasing, mathematical notation, and unit representations. For instance, mathematical answers are evaluated considering formula equivalence and unit conversion, while text-based responses account for semantic similarity and paraphrasing.

At the reasoning level, we conduct a granular step-by-step evaluation for each reasoning chain. Every individual reasoning step undergoes scrutiny for logical consistency with preceding steps, factual accuracy against established knowledge, and relevance to the original question. This meticulous evaluation process identifies not only outright errors but also subtle issues such as logical gaps, unsupported assumptions, and irrelevant tangential reasoning. Based on the comprehensive dual assessment results, we categorize all generated samples into three distinct quality tiers. 
\begin{itemize}[leftmargin=*, noitemsep]
\renewcommand\labelitemi{$\diamond$}
    \item \textit{Category A (High Quality)}: Both answer and reasoning process are correct.
    \item \textit{Category B (Moderate Quality)}: Correct final answers with reasoning process issues.
    \item \textit{Category C (Low Quality)}: Incorrect answers or severely flawed reasoning, automatically discarded.
\end{itemize}

\textbf{Human-in-the-Loop Quality Enhancement:} For Category B samples and potentially redundant Category A samples, we implement a systematic human-guided refinement process designed to enhance reasoning quality while preserving valuable training data. Our comprehensive human review protocol encompasses several critical enhancement dimensions:

\begin{itemize}[leftmargin=*, noitemsep]
\renewcommand\labelitemi{$\diamond$}
    \item \textit{Category B Sample Refinement:} We focus on correcting and streamlining verbose or redundant reasoning steps to improve logical coherence and conciseness. This involves identifying extraneous reasoning chains, consolidating repetitive logical steps, and enhancing the overall flow of argumentation.
    
    \item \textit{Borderline Category A Sample Enhancement:} We systematically address samples identified with intermediate redundancy scores during the automated assessment pipeline—those falling below the automatic removal threshold yet still exhibiting suboptimal reasoning patterns.
\end{itemize}

This human-in-the-loop approach ensures that samples falling into intermediate quality categories undergo systematic improvement rather than wholesale discarding. This methodology strikes an optimal balance between data preservation and quality assurance, thereby enhancing the overall effectiveness of our reasoning dataset for downstream model training.

\textbf{Dynamic Quality Scoring and Data Utilization Strategy:}
To optimize data utilization, we implement a comprehensive five-point quality scoring system that evaluates samples across multiple dimensions:

\begin{itemize}[leftmargin=*, noitemsep]
\renewcommand\labelitemi{$\diamond$}
    \item \textbf{Score 1 (Poor)}: Simple or ambiguous questions answerable without visual input.
    \item \textbf{Score 2 (Below Average)}: Questions with obvious answers or excessive reliance on common sense.
    \item \textbf{Score 3 (Average)}: Clear questions requiring basic image understanding but minimal reasoning
    \item \textbf{Score 4 (Good)}: Questions demanding reasoning about spatial relationships, etc.
    \item \textbf{Score 5 (Excellent)}: Highly multimodal-dependent questions requiring advanced reasoning such as causal inference, occlusion reasoning, or detailed attribute analysis
\end{itemize}

Based on these quality scores, we implement an adaptive data utilization strategy where higher-quality samples are used more frequently during training. Specifically, samples scoring four or five points are repeated multiple times in the training dataset to reinforce high-quality reasoning patterns, while lower-scoring samples are used sparingly to avoid reinforcing suboptimal behaviors. This strategic approach ensures that the model's learning process is dominated by the most valuable and challenging examples while maintaining overall dataset diversity.

The entire automated pipeline demonstrates remarkable efficiency and consistency, processing large volumes of input data while maintaining stringent quality standards across diverse domains and task types. The systematic integration of automated generation, rigorous quality assessment, targeted human enhancement, and adaptive utilization creates a comprehensive framework for producing high-quality training data suitable for effective multimodal model cold-start scenarios.

\subsubsection{Model Merging with Domain Specific Experts}
We conduct a comprehensive analysis of the LongCoT cold start model's performance across various benchmarks using the aforementioned training data, with the objective of identifying and addressing model deficiencies prior to the RL phase. Our analysis reveals concentrated weaknesses in three primary domains: pure text processing, mathematical reasoning, and OCR. To address these limitations, we develop a systematic approach involving specialized data collection and expert model training, followed by model merging to enhance Keye-VL-1.5's foundational capabilities.


\textbf{OCR Capability Enhancement:} Beyond standard OCR datasets, we address specific weaknesses in specialized recognition tasks including license plates, street signage, and official seals. Our enhancement strategy involves three key components: First, we systematically gather OCR datasets targeting identified weak areas, ensuring annotation accuracy through rigorous quality control processes. Second, we develop an automated data pipeline that utilizes images paired with verified OCR annotations to generate relevant OCR questions through other MLLMs, with original annotations serving as ground truth answers to guarantee correctness. Finally, we conduct SFT on the cold-start model using both general-purpose OCR data and our specialized weak-area datasets to create an OCR expert model.


\textbf{Model Merging:} 
We employ model merging  \citep{merge,merge2} to integrate domain-specific expert models and the LongCoT cold start model into a general model for enhanced performance. 

\subsection{Iterative General RL}

Based on the cold-start model, we design our General RL process to further enhance Keye-VL-1.5's reasoning ability, which applies the GSPO \citep{gspo} (Group Sequence Policy Optimization) algorithm for RLVR (Reinforcement Learning with Verifiable Rewards) training, and employs a cyclical iterative approach to collaboratively enhance both the RL model and the cold-start model.

\subsubsection{General RLVR Training}
\textbf{Training data:} 
We select data from domains including mathematics, science \& technology problem, logical reasoning \& puzzle problems, code, chart question answering, visual grounding, spatial relationships, and counting to construct the RLVR training set. Each data point contains a verifiable answer used for rule-based reward calculation. We sample data from different domains according to ablation experiments, analyzing the impact of domain-specific data on model metrics. We then increase the proportion of data from domains that contribute positively to performance improvements. 

\textbf{Training Algorithm:} 
Based on sequence-level importance weight, GSPO employs the following sequence-level optimization objective:
\begin{equation}
\mathcal{J}_{\text{GSPO}}(\theta) = \mathbb{E}_{x \sim \mathcal{D}, \{y_i\}_{i=1}^{G} \sim \pi_{\theta_{\text{old}}}(\cdot | x)} \left[ \frac{1}{G} \sum_{i=1}^{G} \min \left( s_i(\theta) \hat{A}_i, \text{clip}(s_i(\theta), 1 - \epsilon, 1 + \epsilon) \hat{A}_i \right) \right]
\end{equation}

where the group-based advantage estimation is defined as:
\begin{equation}
\hat{A}_i = r(x, y_i) - \text{mean} \left( \{r(x, y_i)\}_{i=1}^{G} \right), \quad \text{std} \left( \{r(x, y_i)\}_{i=1}^{G} \right)
\end{equation}
and  the importance ratio based on sequence likelihood $s_i(\theta)$ is defined as: 
\begin{equation}
s_i(\theta) = \frac{\pi_{\theta}(y_i | x)}{\pi_{\theta_{\text{old}}}(y_i | x)} \quad \text{where} \quad s_i(\theta) = \exp \left( \frac{1}{|y_i|} \sum_{t=1}^{|y_i|} \log \left( \frac{\pi_{\theta}(y_{i,t} | x, y_{i,<t})}{\pi_{\theta_{\text{old}}}(y_{i,t} | x, y_{i,<t})} \right) \right)
\end{equation}

\subsubsection{Progressive Hint Sampling}
During the training process, we find that the model struggles to generate correct responses for some difficult samples, reflecting a deficiency in the model's capabilities. To make full use of these challenging samples and enhance Keye-VL-1.5's reasoning ability, we apply the progressive hint sampling method to improve the success rate of sampling difficult samples.

We first identify the hard cases in the RLVR dataset where Keye-VL-1.5 consistently fails across multiple attempts, then select data with reliable reference answers, sufficient difficulty, and appropriate challenge level as samples for progressive hint sampling.

Unlike the approach of partitioning hints by step, we follow the Minimal Intervention principle to design a hierarchical hint system, aiming to provide the model with the minimal information necessary to solve the problem. We divide the hints into five levels, from abstract concepts to specific reasoning steps:

\begin{itemize}[leftmargin=*, noitemsep]
\renewcommand\labelitemi{$\diamond$}
    \item \textbf{Level 1 (Concept / Observation):} Guide the model to focus on the core concept of the problem or the key features of the image. This level should not contain any problem-solving methods or formulas.
    
    \item \textbf{Level 2 (Strategy / Method):} Suggest one or more possible problem-solving strategies or approaches. For example, "Think holistically", "Try discussing by cases", or "Establish a coordinate system". This level should not mention specific formulas or calculation steps.
    
    \item \textbf{Level 3 (Tools / Formula):} Provide hints for specific mathematical theorems, formulas, or tools needed to solve the problem. For example, "You may need to use the Pythagorean theorem" or "Consider using integration". This level should not provide specific calculation steps.
    
    \item \textbf{Level 4 (Steps / Calculation):} Provide the first concrete operational step in the problem-solving process.
    
    \item \textbf{Level 5 (Complete Solution):} Provide a complete and clear final solution, a perfect solution that can be used as a standard answer.
\end{itemize}

For each hard case, we place the hint information after the query and progressively provide hints from low level to high level. When Keye-VL-1.5 can generate correct response based on a particular level of hint, we consider the hint at that level as the minimal information required to help Keye-VL-1.5 solve the hard case. The responses generated based on this minimal information is then applied to update the policy. In \Cref{tab:partial_solution}, we report the impact of different levels of hints on the sampling success rate of Keye-VL-1.5 in hard cases, aiming to demonstrate the rationality of our hierarchical hint system and the effectiveness of hints in improving the utilization efficiency of hard cases.

\subsubsection{Iterative General RL \& Cold-Start Enhancement}
To improve the learning efficiency on reasoning data and break through the performance bottleneck of the SFT model, we design a multi-round iterative paradigm that collaboratively enhances both the cold-start model, which serves as the starting point for General RL, and the model after General RL.

Our iterative pipeline is as follows:

\begin{enumerate}[leftmargin=*]
    \item Apply the cold-start model as the initial model and perform General RL training.
    \item Apply the model after General RL for rejection sampling on the Cold Start dataset, score the samples with our reward model. If the sampled results are better than the ground truth, update that data point by replacing the ground truth with the sampled results.
    \item Take the updated cold-start data to train a new cold-start model, which serves as the initial model for the next round of General RL.
    \item Take the updated cold-start model to filter the General RL dataset, selecting data with sampling accuracy between 0 and 1 for next round General RL training.
\end{enumerate}

\subsection{Alignment RL}

After General RL, we perform Alignment RL to comprehensively improve the Keye-VL-1.5's performance in real-world application scenarios. We have developed a diversified task system and reward modeling framework to enhance the model's capabilities in the following dimensions:
\begin{itemize}[leftmargin=*, noitemsep]
\renewcommand\labelitemi{$\diamond$}
    \item \textbf{Instruction Following}: Improve the model's ability to generate responses that meet user requirements in terms of content, format, length, and structured output.
    \item \textbf{Format Adherence}: Ensure that the model's responses conform to predefined formats, such as think-answer, agentic think, auto-think, and no-think.
    \item \textbf{Preference Alignment}: For open-ended questions, enhance the reliability, interactivity, and style of the model's responses to improve user experience.
\end{itemize}

\subsubsection{Reward System Design}
The reward system we employ is composed of three main categories:

\begin{itemize}[leftmargin=*, noitemsep]
\renewcommand\labelitemi{$\diamond$}
    \item \textbf{Rule-Based Reward}: Rule-Based reward checks whether the model response adheres to predefined structural and formatting rules, including logical reasoning format (such as think/no\_think/auto\_think formats), as well as structure-specific guidelines such as json, markdown, and code formatting.
    
    \item \textbf{Generative Reward}: For data with ground truth that can not be easily evaluated by rules, we design instructions to prompt MLLMs to access model’s response based on how well it aligns with the reference, its reasoning consistency, and the relatedness to key attributes. Additionally, for ``security and ethics'' tasks, instructions are designed to evaluate whether the responses contain politically errors, misinformation, or offensive content.
    
    \item \textbf{Model-Based Reward}: For tasks without ground truth, the model's responses are scored based on our reward model. Our reward model evaluates whether the responses align with human preferences, promoting responses that adhere to ethical standards.
\end{itemize}

This reward system helps guide the model towards producing accurate, ethical, and contextually appropriate outputs across various tasks.

\subsubsection{Data Construction}

For instruction-following task, we design 25 types of hard constraints, including ``keywords inclusion,'' ``punctuation,'' ``pronunciation,'' ``output format,'' etc., as well as 20 types of soft constraints, such as text style and semantics. We construct a query set consisting of 17k multimodal data and 23k pure text data, with each query assigned 2 to 6 types of constraints as inputs. Hard and soft constraints are rewarded through rule-based rewards and generative rewards, respectively.


For reasoning task, we construct 12k mathematical and logical reasoning queries, with 3 to 5 problem-solving steps designed for each query. The model is required to solve the problem following the prescribed steps. We use rule-based rewards to calculate the correctness of the outcome, and generative rewards to assess whether the reasoning process follows the predefined steps.

For RAG task, we collect a series of instances based on the latest news that require internet searches to obtain answers. We encourage the model to use search and summary behaviors during the think process, ultimately generating the correct answer. We take generative rewards to evaluate the effectiveness of the search behavior in resolving the query, the correctness of the summary behavior, and the consistency of the final answer. We still take GSPO algorithm to optimize our model during Alignment RL.
\section{Training Infrastructure}

To efficiently train MLLMs, we make in-depth infrastructure optimization to address three major challenges: architectural heterogeneity, load imbalance, and I/O bottlenecks.

\textbf{Heterogeneous Hybrid Parallel Strategy}: The training bottleneck of MLLMs stems from computational imbalance caused by architectural heterogeneity. The computational characteristics and resource demands of ViT and LLM are vastly different, and unified parallel strategy leads to significant resource wastage. To address this, we design a heterogeneous hybrid parallel strategy: for the relatively fixed computational pattern of the ViT component, we only use data parallelism (DP) to maximize throughput; whereas for the highly parameter- and memory-intensive LLM, we adopt a hybrid parallelism strategy that combines pipeline (PP), tensor (TP), and data parallelism (DP). This refined strategy is a decisive technical prerequisite for achieving 128K ultra-long sequence training of Keye-VL-1.5.

\textbf{Dynamic Load Balancing Mechanism}: Multimodal data inherently leads to load imbalance, primarily due to the correlation between computational load in the visual encoding phase and the input samples. For instance, processing a high-resolution video incurs significantly more computational cost than a static image. In data parallel training, this leads to GPUs processing complex visual input consumes a longer time while other GPUs finish earlier and waits. To address this, we pre-estimate the time complexity of each sample and then use a greedy algorithm to allocate the samples across different GPUs, thereby balancing the total step duration across all GPUs and improving overall hardware utilization.

\textbf{Flexible and Scalable Dataloader}: To fundamentally resolve I/O bottlenecks, we design a flexible and scalable dataloader that deeply senses the topology of parallel training. In terms of data parallelism (DP), each process only loads a shard of the global dataset; in terms of pipeline parallelism (PP), only the first stage (PP0) is responsible for data acquisition and preprocessing; and in tensor parallelism (TP/CP), the data is first fetched by a single process within the group and efficiently broad-casted across processes. Furthermore, we implement an I/O server architecture to offload CPU-intensive tasks such as video decoding from the training nodes, effectively resolving CPU bottlenecks caused by complex media processing. Finally, we implement a instance-level perfect resume mechanism, ensuring that tasks can seamlessly resume from the last successfully processed sample after an interruption, significantly improving the stability and efficiency of large-scale training.
\begin{table}[t]
\centering
\footnotesize
\begin{threeparttable}
\setlength{\tabcolsep}{1.5mm}{
\resizebox{0.99\textwidth}{!}{
\begin{tabular}{lcccccc} \toprule
    Models & ImageNet-1K & ImageNet-V2 & ImageNet-A & ImageNet-R & ImageNet-S & ObjectNet \\ 
    \midrule
    Base (SigLIP-400M-384-14) & \textbf{83.08} & \textbf{77.34} & \underline{82.22}& \textbf{95.78}& \textbf{74.59} &\underline{76.99} \\
     + 1D interpolation & 82.02	&75.96	&80.92	&94.50	&70.74	&67.58 \\
     + 1D interpolation + 2D RoPE &  \underline{82.65}	&\underline{76.80}	&\textbf{83.26}	&\underline{95.22}	&\underline{72.59}	&\textbf{78.70} \\
    \midrule
\end{tabular}}
}
\end{threeparttable}
\caption{\textbf{Comparison of ViT variants on the ImageNet benchmarks:} The highest scores are marked in \textbf{bold} and the second highest are \underline{underlined}.}
\label{tab:vitzeroshot}
\end{table}


\begin{figure*}[t]
\centering

\begin{subfigure}[b]{\linewidth}
    \centering
    \includegraphics[width=\linewidth]{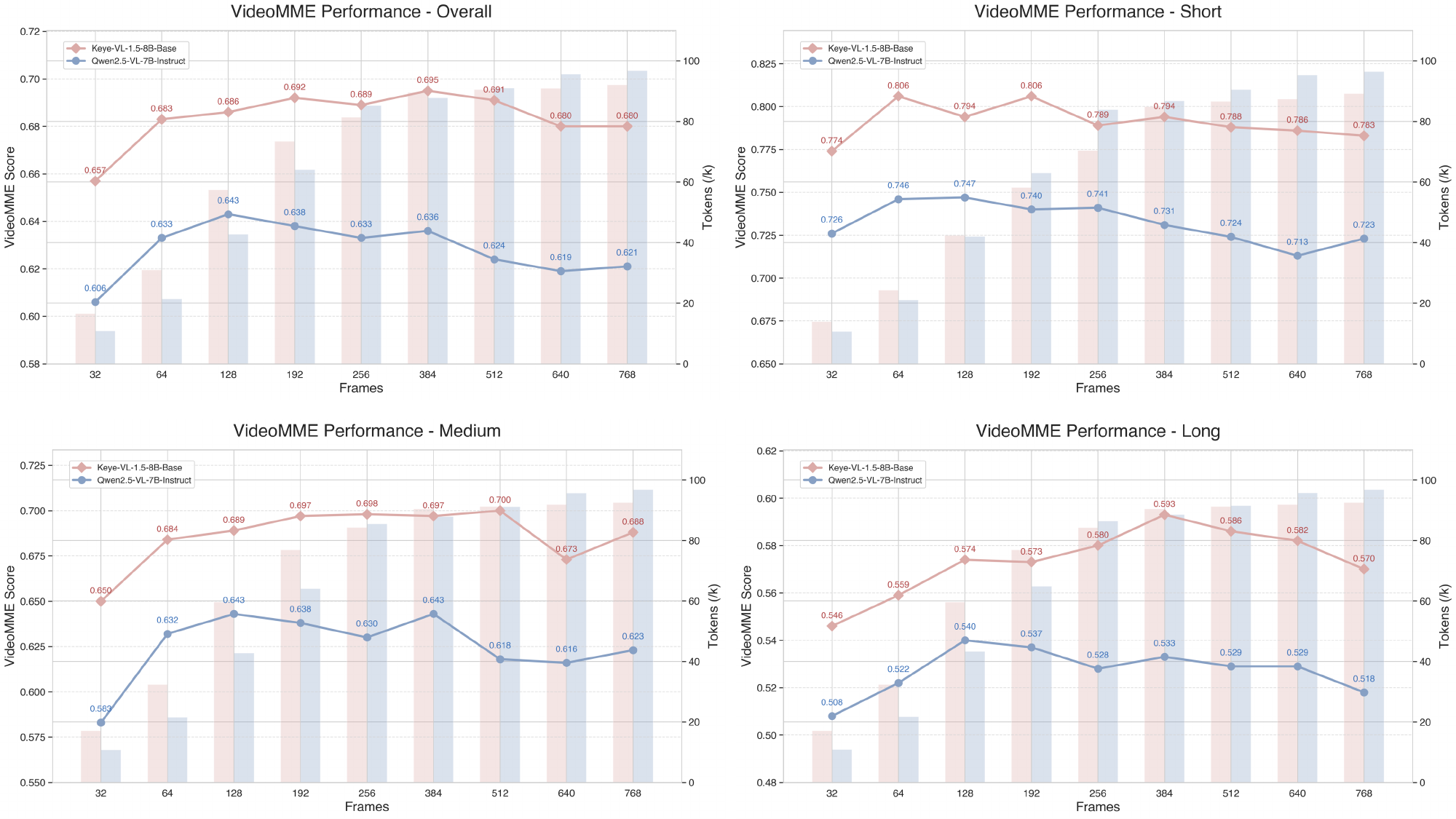}
    \caption{Frames.}
    \label{fig:slowfast_abl_medium}
\end{subfigure}

\begin{subfigure}[b]{\linewidth}
    \centering
    \includegraphics[width=\linewidth]{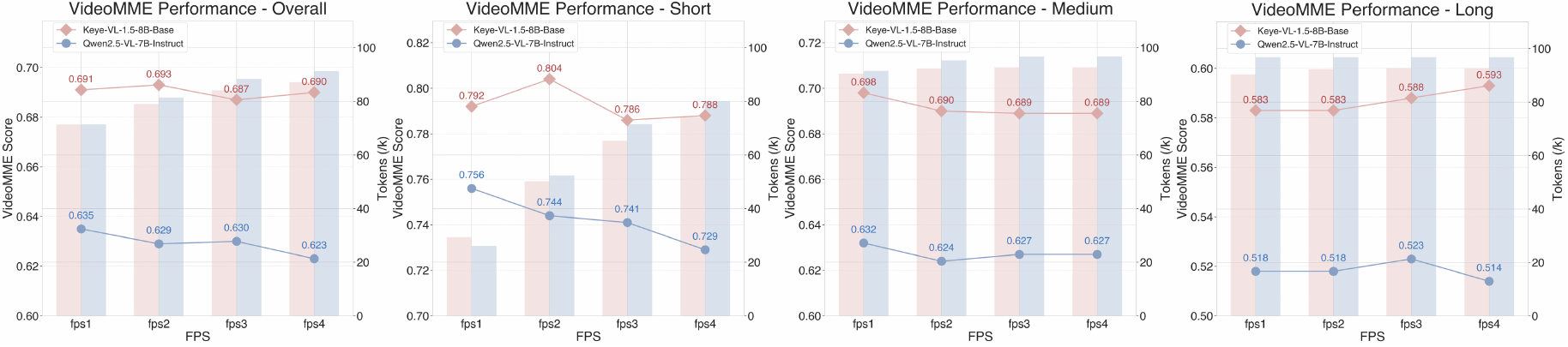}
    \caption{FPS.}
    \label{fig:slowfast_abl_short}
\end{subfigure}

\caption{SlowFast (Keye-VL-1.5-Base) and 2D convolution (Qwen-2.5 VL) video encoding strategies were compared on VideoMME across different video lengths. Keye-VL-1.5-Base exhibits strong visual understanding capabilities across various settings, e.g., diverse frame numbers and FPS.}
\label{fig:slowfast_abl}
\end{figure*}

\section{Evaluation}

\subsection{Zero-shot Image Classification of ViT}
To validate that our continue trained native-resolution ViT is able to capture promising visual representations, we conduct a wide-used zero-shot image classification benchmark analysis.
In our evaluation, we perform a comparative analysis between the base SigLIP model and its two native-resolution position embedding variants, leveraging the CLIP Benchmark\footnote{\url{https://github.com/LAION-AI/CLIP_benchmark}} framework with text prompt template\footnote{\url{https://colab.research.google.com/github/openai/clip/blob/master/notebooks/Prompt_Engineering_for_ImageNet.ipynb\#scrollTo=sRqDoz1Gbsii}}.

The evaluation covers six benchmark datasets: ImageNet-1K, ImageNet-V2, ImageNet-A, ImageNet-R, ImageNet-S and ObjectNet, and its results are shown in Table~\ref{tab:vitzeroshot}.
From it, we have the following observations:
(1) Compared with base SigLIP model, our 1D interpolation position embedding native-resolution model variant has slightly performance degeneration, the reason might be the interpolated 1D position encoding cannot uniquely identify the underlying 2D patch arrangement. For instance, a sequence of 196 patches may correspond to multiple distinct spatial configurations (e.g., 14×14, 7×28, or 28×7), leading to ambiguous spatial localization during feature projection.
(2) With 2D RoPE modification, our ViT could clearly perceive the shape of the image, and showing competitive results with Base SigLIP performance (the best and runner-up results). We think the reason maybe our continued pretraining corpus sharing the same distribution with our MLLMs, rather than the Image-Text matching task.

\subsection{SlowFast Video Encoding Strategy Discussion}
In this section, to verify that our SlowFast strategy can capture fine-grained video information, we conduct a comparative analysis between Keye-VL-1.5-Base and Qwen-2.5-VL. Keye-VL-1.5-Base is a pre-trained model equipped with our SlowFast technique, while Qwen-2.5-VL employs a 2D convolution merging technique for video compression.

For a fair comparison, we evaluate both models on the VideoMME benchmark under different settings. Specifically, we test with fixed frame numbers ranging from 32, 64, 128, up to 768, and FPS values from 1 to 4 in increments of 1.
Meanwhile, different with linear token budget increasing of 2D convolution along with frame amount, our slowFast strategy has highly adaptive token budget for different videos with different information density.
Combining the two factors, we show the prediction performances and the LLM-side visual token budgets across different video category (i.e., short/medium/long and overall) at the  Figure~\ref{fig:slowfast_abl}.
According to it, we have the following observations:
%

\begin{itemize}[noitemsep]
\renewcommand\labelitemi{$\diamond$}
\item  In terms of the overall score of VideoMME, our Keye-VL-1.5-Base shows the similar performance trend as Qwen-2.5-VL. Specifically, both of them show an increase then decline performance trend, while our Keye-VL-1.5-Base achieves its best performance at 384 frames, Qwen reaches its peak at 128 frames, indicating that our SlowFast is also a reliable video encoding strategy.
\item For the sub-category performance of VideoMME, the Qwen-2.5-VL show the inflection point at 128/384/128 and our Keye-VL-1.5-Base shows the inflection point at 192/512/384 for the three categories of short, medium, and long videos. Compared with Qwen-2.5-VL, The fact that Keye’s performance begins to decline at a later point demonstrates that our SlowFast strategy enables the LLM to integrate multi-frame information more effectively.
\item In terms of video token usage on the LLM side, Qwen-2.5-VL demonstrates a nearly linear relationship with the number of frames and token budget. In contrast, our Keye-VL-1.5-Base generates more visual tokens when the frame number is low, but fewer visual tokens when the frame number is high compared to Qwen-2.5-VL. This phenomenon demonstrates that our SlowFast strategy is more flexible and makes more efficient use of computational resources.
\item For the different FPS setting, we could observe that our Keye-VL-1.5-Base is more stable in evaluation results than Qwen-2.5-VL. Additionally, our SlowFast video encoding approach achieves a flexible token budget on par with the 2D convolution technique.
\end{itemize}

\subsection{Public Benchmarks}

\begin{table}[t]
    \centering
    \small
    \renewcommand{\arraystretch}{1.2}
    \resizebox{0.99\textwidth}{!}{
    \begin{tabular}{l  | S[table-format=2.1] c c | S[table-format=2.1] S[table-format=2.1] S[table-format=2.1] | S[table-format=2.1] S[table-format=2.1]}
    \toprule
    \multirow{2}{*}{\textbf{Benchmark}}  & \textbf{Keye-VL-1.5} & \textbf{Keye-VL-Preview} & \multirow{2}{*}{} & \textbf{Qwen2.5-VL} & \textbf{InternVL3} & \textbf{MiMo-VL}  & \multicolumn{1}{c}{\multirow{2}{*}{\textbf{GPT-4o}}} & \multicolumn{1}{c}{{\textbf{Claude 3.7}}} \\
    & \textit{8B-Thinking} & \textit{8B-Thinking} & & \textit{7B} & \textit{8B} & \textit{7B-RL 2508} &   & \textit{Sonnet} \\
    \midrule
    \rowcolor{kuaishoublue!20}
    \multicolumn{9}{l}{\textbf{General}} \\ %
    OpenCompass          & \textbf{79.5}  & {77.4}         &   & 70.9  & 73.6  & \underline{75.2}  &   72.0   & 70.1 \\
    MMMU$_{\mathrm{val}}$          & \textbf{71.4}  & \underline{71.4}         &   & 58.6  & 62.7  & {69.4}  &   70.7   & 69.8 \\
    AI2D                                  &  \textbf{89.5} & {86.7}        &   & 83.9  & {85.2}  & \underline{87.1}  &  82.6 &  81.4 \\
    MMBench                                  &  \textbf{92.0} & \underline{92.0}        &   & 82.2  & {82.1}  & {86.8}  &  86.0 &  79.7 \\
    BLINK$_{\mathrm{val}}$                &  54.9 & 52.0        &   & \underline{56.4}  & 55.5  & \textbf{62.2} &  60.0  &  62.3 \\
    ZeroBench$_{\mathrm{sub}}$            & \underline{16.2} & {15.2}         &   & {0.0} & {0.0} & \textbf{18.2} & {-} & {-} \\
    VisuLogic                             & {23.1} & \underline{25.6}         &   & {20.0} & \textbf{26.1} & {24.5} & {-} & {-} \\
    RealWorldQA                           & \textbf{73.5} & {67.7}         &   & {68.2} & {70.6} & \underline{71.0} & {-} & {-} \\
    SimpleVQA                             & \underline{42.9} & 41.6         &   & {41.4} & {35.1} & \textbf{44.9} & {-} & {-} \\
    MMStar                                & \textbf{80.5} & \underline{75.5}         &   & {64.9} & {68.4} & {73.7} & {-} & {-} \\
    MMVP                                  & \underline{80.7} & {79.0}         &   & {78.0} & {78.3} & \textbf{81.7} & {-} & {-} \\
    HallusionBench                        & {62.7} & \textbf{67.0}         &   & {55.7} & {49.4} & \underline{65.2} & {-} & {-} \\
    OCRBench                              & 86.6 & 85.1        &   & \textbf{89.7} & \underline{88.0}  &   82.2    & 84.3 &  80.6 \\
    \rowcolor{kuaishoublue!20}
    \midrule
    \multicolumn{9}{l}{\textbf{Video}} \\ %
    Video-MME$_{\mathrm{w/o~sub.}}$       & \textbf{73.0} & 67.7         &   & 65.1 & 66.3  &    \underline{68.9}   & 71.9    &     {-} \\
    Video-MMMU                            &  \textbf{66.0}   & {57.6}         &   & 47.4  &   {48.9} &  \underline{59.5}    & {-}    &    {-} \\
    TempCompass            & \textbf{75.5}  & 71.5         &   & 68.3 &   \underline{70.8} & {-}   & {-}    &    {-} \\
    LongVideoBench                          &   \textbf{66.0} &  {62.8}         &   & 59.3  &   {63.9}    &    \underline{64.9}   & {-}    &  {-} \\
    MMVU                          &   \textbf{68.3} & \underline{66.3}        &   & {45.5}  &   {39.4}    &    {-}   & {-}    &  {-} \\
    \midrule
    \rowcolor{kuaishoublue!20}
    \multicolumn{9}{l}{\textbf{MATH}} \\ %
    MathVision                           & \underline{46.8} & 46.0       &   & {26.2} & {28.8} & \textbf{48.7} & {31.2} & {-} \\
    MathVista$_{\mathrm{MINI}}$          & \textbf{81.2} & 80.7        &   & {66.8} & {70.7} & \underline{79.0} & {63.8} & {-} \\
    MathVerse$_{\mathrm{vision}}$          & \underline{68.7} & 59.8        &   & {44.9} & {32.4} & \textbf{74.8} & {49.9} & {-} \\
    OlympiadBench                        & {47.5} & \underline{54.8}        &   & {19.4} & {25.9} & \textbf{56.4} & {25.9} & {-} \\
    WeMath          & \textbf{67.5} & 60.7       &   & {37.7} & {38.5} & \underline{65.2} & {50.6} & {-} \\
    LogicVista          & \underline{58.8} & 54.8        &   & {44.5} & {43.6} & \textbf{63.5} & {54.4} & {-} \\
    DynaMath          & \underline{39.7} & 37.3       &   & {20.1} & {23.9} & \textbf{48.7} & {54.4} & {-} \\
    \bottomrule
    \end{tabular}
    }
    \caption{
        \textbf{Comparison of Keye-VL-1.5 in \textit{Thinking} mode with Keye-VL-Preview and other models on diverse visual-language benchmarks:} The best results among open-source models are \textbf{bolded} and the second-best results are \underline{underlined}. 
    }
    \label{tab:general_updated}
\end{table}

In this section, we evaluate Keye-VL-1.5 across various benchmarks. For \textit{general vision-language tasks}, we select $\text{OpenCompass}$ (\cite{opencompass}), $\text{MMMU}$ (\cite{mmmu}), AI2D (\cite{ai2d}), $\text{MMBench}$ (\cite{mmbench}), $\text{BLINK}$ (\cite{blink}), ZeroBench (\cite{zerobench}), VisuLogic (\cite{visu_logic}), RealWorldQA (\cite{real_world_qa}), SimpleVQA (\cite{simple_vqa}), MMStar (\cite{mm_star}), MMVP (\cite{mmvp}), HallusionBench (\cite{hallusionbench}) and OCRBench (\cite{ocrbench}). For \textit{public Video tasks}, we select Video-MME(\cite{video_mme}), Video-MMMU (\cite{video_mmmu}), TempCompass (\cite{tempcompass}), LongVideoBench (\cite{longvideobench}), and MMVU (\cite{mmvu}). For \textit{MATH tasks}, we select MathVision (\cite{mathvision}), MathVista\textsubscript{MINI} (\cite{mathvista}), MathVerse\textsubscript{vision} (\cite{mathverse}), OlympiadBench (\cite{olympiadbench}), WeMath (\cite{wemath}), LogicVista (\cite{logicvista}), and DynaMath (\cite{dynamath}). 

 We compare the performance of Keye-VL-1.5 in \textit{Thinking} mode with Keye-VL-Preview and other state-of-the-art models of a similar scale, including Qwen2.5-VL 7B, InternVL3-8B (\cite{internvl3}), MiMo-VL-7B-RL 2508 (\cite{mimovl}), and proprietary models such as GPT-4o and Claude-3.7-Sonnet.

On general vision-language tasks, Keye-VL-1.5 demonstrates competitive performance across most benchmarks, often achieving SOTA or near SOTA results and outperforming other models overall. On the large-scale general benchmarks OpenCompass, $\text{MMMU}_\text{val}$ and AI2D, Keye-VL-1.5 obtains scores of 79.5\% 71.4\% and 86.7\% respectively, surpassing all other models. On MMBench and MMStar, Keye-VL also achieves the best performance. In mathematical reasoning tasks, Keye-VL-1.5 significantly outperforms Qwen2.5-VL 8B and InternVL3-8B, achieving comparable results with MiMo-VL 7B-RL.

In video-centric scenarios, Keye-VL-1.5 demonstrates superior capabilities compared to other open-source models. Our evaluations indicate that an accurate understanding of video content is Keye-VL-1.5's core advantage. On public video benchmarks, Keye-VL-1.5  significantly outperforms other models, particularly on Video-MMMU, with an absolute improvement of 6.5\%.

\subsection{Internal Benchmarks}
Despite extensive evaluations on a wide array of public video benchmarks, these benchmarks exhibit numerous limitations that necessitate a focused effort on developing a proprietary, internal evaluation suite. The primary issues are as follows:

\begin{itemize}[leftmargin=*, noitemsep]
\renewcommand\labelitemi{$\diamond$}
    \item \textbf{Limited Task Coverage}: Current publicly available benchmarks primarily focus on basic perception and simple reasoning capabilities, with insufficient coverage of specialized domains and temporal understanding tasks, failing to comprehensively evaluate model performance across diverse scenarios.
    
    \item \textbf{Oversimplified Question Formats}: Existing evaluation tasks tend to employ overly simplistic questioning approaches. For instance, in video question answering tasks, queries often involve only the most basic inquiries about video content, such as counting the number of people present, which inadequately reflects real-world complexity.
    
    \item \textbf{Restrictive Answer Methodologies}: To facilitate accuracy computation, questions are typically abstracted into yes/no responses or multiple-choice formats, which significantly deviate from natural user interaction patterns and limit the assessment of model's genuine conversational capabilities.
    
    \item \textbf{Data Contamination Risks}: Since datasets are publicly available, there exists a non-negligible possibility that models have already encountered these data during training, potentially leading to inflated performance metrics and compromised evaluation validity.
    
    \item \textbf{Language and Cultural Bias}: Many existing test sets exhibit bias toward English-language scenarios, limiting our understanding of model performance in Chinese usage contexts and failing to capture culture-specific nuances and requirements.
\end{itemize}

Therefore, we construct a rigorous internal video evaluation benchmark. The video sources include both internal and external platform content, as well as artificially constructed videos, with resolutions ranging from 360p to 1440p, effectively avoiding overlap with existing training data. The questions are categorized into several dimensions to provide comprehensive coverage: Visual Element Recognition for assessing visual element identification capabilities, Reasoning Ability for evaluating logical reasoning skills, Temporal Info Understanding for measuring temporal information comprehension, Knowledge-based QA for testing knowledge-grounded question answering, Description Ability for evaluating descriptive capabilities, Robustness for testing model stability, Creative Ability for assessing creative thinking, and Domain Expertise for evaluating specialized domain knowledge.

The scoring methodology employs comparative evaluation across multiple model results and GSB (Good, Same, Bad) preference selection. The baseline models can be either GPT-4o or Gemini 1.5 Pro. The specific evaluation approach involves two methods. First, the scoring method uses multiple models (typically 2) to generate results that are evaluated separately on a 1-5 scale. Three annotators score the answers based on the video content and reference annotation guidelines, providing both fine-grained and overall scores. Second, the GSB method involves direct comparison between two model results using Good-Same-Bad preference selection. When two answers have significantly different scores, the higher-scoring answer is preferred. When the scores are similar, the selection is based on annotation rules and subjective judgment to determine which answer is better. If no clear distinction can be made, the selection reflects whether both answers are equally good, equally poor, or equally average based on answer quality.

\begin{table}[t]
\centering
\footnotesize
\begin{threeparttable}
\setlength{\tabcolsep}{1.5mm}{
\resizebox{0.99\textwidth}{!}{
\begin{tabular}{l|cccccc} \toprule
\textbf{Model} & \textbf{Average} & \textbf{Correctness} & \textbf{Completeness} & \textbf{Relevance} & \textbf{Fluency} & \textbf{Creativity} \\
\midrule
Keye-VL-1.5-8B & \textbf{3.53} & \textbf{3.73} & 4.62 & 4.85 & 4.59 & 3.64 \\
MiMoVL-7B-RL-2508 & 3.40 & 3.54 & \textbf{4.63} & \textbf{4.93} & \textbf{4.82} & \textbf{3.79} \\
\midrule
\multicolumn{7}{l}{\textbf{Performance Comparison:}} \\
vs. MiMoVL-7B-RL-2508 & +0.13 & +0.19 & -0.01 & -0.08 & -0.23 & -0.15 \\
vs. Keye-VL-Preview & +0.51 & +0.57 & +0.25 & +0.11 & -0.24 & -0.26 \\
\bottomrule
\end{tabular}}
}
\end{threeparttable}
\caption{\textbf{Comprehensive capability evaluation comparison:} This table presents the performance comparison between Keye-VL-1.5-8B and MiMoVL-7B-RL-2508 across multiple dimensions including correctness, completeness, relevance, fluency, and creativity. Performance differences against baseline models are also provided, with the highest scores marked in \textbf{bold}. Positive values indicate performance improvements, while negative values indicate performance degradation.}
\label{tab:multimodal_eval}
\end{table}%

\begin{table}[t]
\centering
\footnotesize
\begin{threeparttable}
\setlength{\tabcolsep}{1.2mm}{
\resizebox{0.99\textwidth}{!}{
\begin{tabular}{l|ccccccccc} \toprule
\textbf{Model Version} & \makecell{\textbf{Visual Element}\\ \textbf{Recognition}} & \makecell{\textbf{Reasoning}\\ \textbf{Ability}} & \makecell{\textbf{Temporal Info}\\ \textbf{Understanding}} & \makecell{\textbf{Knowledge-based}\\ \textbf{QA}} & \makecell{\textbf{Description}\\ \textbf{Ability}} & \textbf{Robustness} & \makecell{\textbf{Creative}\\ \textbf{Ability}} & \makecell{\textbf{Domain}\\ \textbf{Expertise}} & \textbf{Overall} \\
\midrule
\textit{Number of Cases} & 35 & 27 & 22 & 30 & 11 & 24 & 29 & 22 & 200 \\
\midrule
Keye-VL-1.5-8B & 3.49 & \textbf{3.81} & \textbf{3.36} & 2.50 & 3.73 & \textbf{4.29} & 3.66 & 3.68 & \textbf{3.53} \\
MiMoVL-7B-RL-2508 & \textbf{3.49} & 3.56 & 3.18 & \textbf{2.60} & \textbf{3.91} & 3.46 & \textbf{3.66} & \textbf{3.68} & 3.40 \\
\midrule
\multicolumn{10}{l}{\textbf{Performance Comparison:}} \\
vs. MiMoVL-7B-RL-2508 & 0.00 & +0.25 & +0.18 & -0.10 & -0.18 & +0.83 & 0.00 & 0.00 & +0.13 \\
vs. Keye-VL-Preview & +0.35 & +1.00 & +0.77 & +0.27 & +0.46 & +0.41 & +0.11 & +0.91 & +0.51 \\
\bottomrule
\end{tabular}}
}
\end{threeparttable}
\caption{\textbf{Detailed capability evaluation across multiple dimensions:} This table presents a comprehensive comparison of Keye-VL-1.5-8B and MiMoVL-7B-RL-2508 across eight core capabilities including visual element recognition, reasoning ability, temporal information understanding, knowledge-based QA, description ability, robustness, creative ability, and domain expertise. The evaluation is based on 200 test cases distributed across different capability categories. The highest scores are marked in \textbf{bold}, and performance differences are provided for comparative analysis.}
\label{tab:detailed_capability_eval}
\end{table}

\subsection{Evaluation Results}

\textbf{Keye-VL-1.5-8B achieves significant performance improvements over previous versions:} As demonstrated in Table~\ref{tab:multimodal_eval}, Keye-VL-1.5-8B establishes a substantial lead with an overall composite score of \textit{3.53}, representing a remarkable \textit{+0.51} improvement over Keye-VL-Preview. This advancement is particularly pronounced in correctness (\textit{+0.57}) and completeness (\textit{+0.25}), demonstrating the model's enhanced ability to provide accurate and comprehensive responses. The model also shows notable gains in relevance (\textit{+0.11}), indicating improved alignment between responses and user queries.

\textbf{The model demonstrates competitive performance against industry benchmarks:} In direct comparison with MiMoVL-7B-RL-2508, Keye-VL-1.5-8B achieves a higher overall score (\textit{3.53 vs. 3.40}), establishing a \textit{+0.13} advantage in composite performance. The model particularly excels in correctness (\textit{+0.19}) while maintaining competitive performance in completeness (\textit{-0.01}). However, the evaluation reveals trade-offs in certain dimensions, with MiMoVL-7B-RL-2508 showing superior performance in fluency (\textit{+0.23}), relevance (\textit{+0.08}), and creativity (\textit{+0.15}). This performance profile indicates that while our model achieves stronger factual accuracy, it faces challenges in language generation sophistication.

\textbf{Detailed capability analysis reveals domain-specific strengths and optimization priorities:} The fine-grained evaluation in Table~\ref{tab:detailed_capability_eval} demonstrates Keye-VL-1.5-8B's exceptional performance across multiple core capabilities. The model achieves decisive advantages in \textit{Reasoning Ability (3.81)}, \textit{Temporal Information Understanding (3.36)}, and \textit{Robustness (4.29)}, with the latter representing a substantial \textit{+0.83} lead over MiMoVL-7B-RL-2508. These results highlight the model's particular strength in handling complex analytical tasks and maintaining consistent performance under challenging conditions. The model matches MiMoVL-7B-RL-2508 in \textit{Visual Element Recognition (3.49)} and \textit{Creative Ability (3.66)}.

\textbf{The model establishes a strong foundation in fundamental visual understanding capabilities:} Keye-VL-1.5-8B's performance demonstrates significant improvements in core visual processing tasks compared to previous iterations. The \textit{+0.35} advancement in visual element recognition and \textit{+1.00} improvement in reasoning ability over Keye-VL-Preview indicate substantial progress in fundamental perceptual and cognitive pathways. Particularly notable is the model's \textit{+0.77} improvement in temporal information understanding, reflecting enhanced capability in processing sequential visual information and understanding dynamic relationships within video content. These foundational improvements provide a robust platform for handling complex multimodal reasoning tasks.


\subsection{Ablation Studies and Findings}

\subsubsection{Effects of SFT, MPO, and Long CoT Cold Start}

Table~\ref{tab:training_comparison} presents a comprehensive evaluation of different training methodologies using varying quantities of high-quality data for SFT and MPO. The experimental results demonstrate that increasing the volume of SFT training data consistently enhances model performance across mathematical reasoning, logical inference, and OCR capabilities. Notably, our carefully curated preference dataset for MPO consistently yields additional performance improvements across all evaluated benchmarks. The implementation of Long CoT cold start training produces particularly remarkable results, with substantial performance gains observed across all benchmarks, most notably in mathematical reasoning tasks. These findings empirically validate the effectiveness of our proposed data processing pipeline and training methodology, demonstrating the synergistic benefits of combining high-quality supervised fine-tuning with preference optimization and strategic initialization approaches.

\begin{table}[]
\caption{\textbf{Performance comparison of different training strategies across multiple benchmarks.} The table shows evaluation results for SFT and MPO training with varying dataset sizes (15k and 128k samples), as well as the Long CoT Cold Start and RL approaches.}\label{tab:training_comparison}

\resizebox{\textwidth}{!}{%
\begin{tabular}{lcccccccccc}
\toprule 
\multicolumn{1}{c}{\textbf{Model}} & \textbf{OpenCompass} & \textbf{MMBCN} & \textbf{MMBEN} & \textbf{MMVet} & \textbf{AI2D} & \textbf{Hallusion} & \textbf{MathVista} & \textbf{MMMU} & \textbf{MMStar} & \textbf{OCR} \\\rowcolor{blue4!40} \midrule
\multicolumn{11}{c}{Baselines} \\
Qwen2.5-VL 7B & 70.56 & 82.66 & 83.28 & 65.60 & 84.39 & 55.97 & 66.60 & 56.56 & 64.60 & 87.80 \\
MiMO-VL-7B & 75.62 & 81.50 & 83.13 & 77.52 & 83.78 & 61.95 & 80.30 & 65.22 & 70.80 & 83.10 \\
Keye-VL-7B-Preview & 77.43 & 90.71 & 92.03 & 68.62 & 87.18 & 61.98 & 78.70 & 71.67 & 75.00 & 84.90 \\ \rowcolor{blue4!30} \midrule
\multicolumn{11}{c}{\textit{SFT+MPO}} \\
SFT-15k & 67.24 & 80.96 & 83.75 & 59.13 & 81.44 & 51.20 & 63.50 & 56.44 & 61.13 & 82.70 \\
MPO-15k & 69.31 & 80.65 & 83.28 & 62.02 & 83.19 & 52.79 & 67.00 & 61.22 & 63.07 & 83.20 \\
SFT-128k & 67.80 & 80.42 & 83.67 & 54.82 & 82.55 & 51.44 & 65.70 & 57.44 & 61.80 & 86.60 \\
MPO-128k & 70.34 & 81.27 & 83.44 & 62.34 & 84.52 & 55.76 & 68.40 & 58.33 & 65.33 & 85.70 \\ \midrule \rowcolor{blue4!20} 
\multicolumn{11}{c}{Long CoT Cold Start} \\
Long CoT Cold Start & 75.32 & 88.24 & 88.93 & 62.89 & 86.04 & 61.05 & 76.40 & 68.33 & 73.20 & 86.10 \\
RFT-SFT & 76.33 & 89.16 & 91.02 & 67.29 & 86.43 & 61.52 & 77.60 & 67.78 & 74.20 & 85.70 \\ \midrule\rowcolor{blue4!10}
\multicolumn{11}{c}{RL} \\
Keye-VL-1.5-RL & 79.41 & 92.88 & 92.88 & 71.19 & 90.35 & 65.68 & 81.30 & 69.00 & 79.20 & 85.70 \\
\& Partial Solution & 80.13 & 93.27 & 93.50 & 73.67 & 89.77 & 66.12 & 82.60 & 71.67 & 80.93 & 85.10 \\ \bottomrule
\end{tabular}%
}
\end{table}

\begin{table}[]
\caption{\textbf{Performance evaluation of expert models and model merging techniques on OCR-related benchmarks.} The table compares baseline models with our approach, including base model, OCR expert model, and the merged configuration.}
\label{tab:ocr_expert_comparison}
\centering
\resizebox{0.9\textwidth}{!}{%
\begin{tabular}{ccccccc}
\toprule 
\multirow{2}{*}{\textbf{Model}} & \multirow{2}{*}{\textbf{AVG}} & \textbf{TextVQA} & \textbf{ChartQA} & \textbf{InfographicVQA} & \textbf{DocVQA} & \textbf{OCRBench} \\
 &  & \textit{Test} & \textit{Test} & \textit{Val} & \textit{Val} & \textit{Test}
 \\\rowcolor{blue4!20} \midrule
\multicolumn{7}{c}{\textit{Baseline}} \\
MiMoVL-7B-RL-2508 & 81.41 & 75.57 & 70.00 & 84.93 & 94.35 & 82.20 \\
Keye-VL-8B-Preview & 79.68 & 75.47 & 86.24 & 66.89 & 84.31 & 85.50 \\ \midrule  \rowcolor{red4!20}
\multicolumn{7}{c}{\textit{Ours}} \\
Base Model & 78.25 & 70.45 & 78.08 & 69.85 & 87.18 & 85.70 \\
OCR Expert & 83.65 & 79.36 & 84.76 & 74.54 & 93.21 & 86.40 \\
Merge OCR + Base & 84.51 & 83.40 & 84.88 & 74.26 & 93.33 & 86.70 \\ \bottomrule
\end{tabular}%
}
\end{table}
\subsubsection{Effectiveness of Expert Models and Model Merging}

Table~\ref{tab:ocr_expert_comparison} demonstrates the effectiveness of our expert model approach and model merging technique, using OCR tasks as a representative case study. Our base model initially achieved an average OCR performance of 78.25\%, comparable to the preview version but exhibiting notable deficiencies in specialized domains such as license plate recognition, seal/stamp identification, and street scene text extraction. To address these limitations, we develop a specialized OCR expert model trained on curated domain-specific data. The OCR expert model demonstrates substantial improvements across all evaluated OCR benchmarks, achieving an average score of 83.65\%. Furthermore, the strategic merging of our base model with the OCR expert yields additional performance enhancements, reaching an average score of 84.51\%. This merged configuration significantly surpasses the perceptual capabilities of MiMo-VL, with particularly notable improvements in TextVQA (83.40\% vs. 75.57\%) and ChartQA (84.88\% vs. 70.00\%).

These empirical results validate the effectiveness of our proposed technical approach, demonstrating that domain-specific expert models can be successfully integrated with general-purpose base models to achieve superior performance across specialized tasks while maintaining overall model capabilities. Additionally, our experiments reveal the following findings:

\textit{Limited Training Steps:} Expert models trained with more steps continue improving within their specialized domains. However, merged model performance initially increases with expert training steps, then decreases, indicating an optimal training duration.

\textit{Limited Learning Rate:} Expert models achieve better performance with smaller learning rates, and the corresponding merged models also perform better.

The parameter divergence between expert and general models significantly affects merged model performance. Small divergences limit domain-specific improvements, while large divergences lead to suboptimal merged performance, creating a critical trade-off between specialization and integration.

\begin{table}[]
\caption{\textbf{Performance comparison of alignment reinforcement learning across instruction following and mathematical reasoning benchmarks.} The evaluation includes both multimodal and text-only instruction following tasks, as well as comprehensive mathematical reasoning assessments. Results are presented for both Think and No-Think inference modes.}
\label{tab:alignment_rl_evaluation}
\resizebox{\textwidth}{!}{%
\begin{tabular}{cccccccccc}
\toprule 
\multirow{2}{*}{\textbf{Model Name}} & \multirow{2}{*}{\textbf{Mode}} & \multicolumn{4}{c}{\textbf{Instruction Following}} & \multicolumn{4}{c}{\textbf{Math Reasoning}} \\ \cmidrule{3-6}
 &  & \textit{MIA-Bench} & \textit{MMIFEval} & \textit{IFEval} & \textit{LiveBench} & \textit{WeMath} & \textit{MathVerse} & \textit{MathVision} & \textit{LogicVista} \\ \rowcolor{blue4!20} \midrule
\multicolumn{10}{c}{Baselines} \\
Keye-VL-8B-preview & Think & 87.60 & 56.97 & 65.80 & 59.30 & 60.76 & 59.77 & 46.22 & 54.81 \\
Keye-VL-8B-preview & No-Think & 89.85 & 56.06 & 73.75 & 53.00 & \multicolumn{4}{c}{-} \\ \midrule  \rowcolor{red4!20}
\multicolumn{10}{c}{Ours} \\
Alignment RL & Think & 91.95 & 63.45 & 70.98 & 64.70 & 64.95 & 61.17 & 48.45 & 57.27 \\
Alignment RL & No-Think & 91.06 & 62.87 & 78.37 & 61.70 & \multicolumn{4}{c}{-} \\ \bottomrule
\end{tabular}%
}
\end{table}

\subsubsection{Effectiveness of Alignment Reinforcement Learning}

To validate the effectiveness of our alignment reinforcement learning approach, we conducted comprehensive evaluations starting from the Keye-VL-8B-preview baseline, focusing on instruction following capabilities and mathematical reasoning performance. Our evaluation framework encompasses both multimodal instruction following benchmarks (MIA-Bench and MMIFEval) and text-only instruction following assessments (IFEval and LiveBench). For mathematical reasoning evaluation, we selected four widely adopted benchmarks to ensure comprehensive coverage of mathematical capabilities. As demonstrated in Table~\ref{tab:alignment_rl_evaluation}, our alignment RL approach consistently outperforms the baseline across both inference modes. In the Think mode, substantial improvements are observed across all instruction following benchmarks, with notable gains of 4.35 points on MIA-Bench (91.95\% vs. 87.60\%), 6.48 points on MMIFEval (63.45\% vs. 56.97\%), and 5.40 points on LiveBench (64.70\% vs. 59.30\%). Similarly, in the No-Think mode, the model demonstrates consistent improvements, particularly achieving a 4.62-point enhancement on IFEval (78.37\% vs. 73.75\%). The mathematical reasoning capabilities also exhibit modest but consistent improvements across all evaluated benchmarks, with average gains ranging from 2-4 points. These results empirically validate that our alignment algorithm effectively enhances functional capabilities in instruction following while simultaneously strengthening general reasoning abilities. The consistent performance improvements across diverse evaluation metrics confirm the robustness and effectiveness of our alignment reinforcement learning methodology.

\begin{table}[]
\centering
\caption{\textbf{Effect of different hint levels on model performance across multiple attempts.} The table compares the percentage of completely incorrect data, average score for four attempts, and standard deviation for each level of hint provided.}
\label{tab:partial_solution}
\begin{tabular}{@{}lccc@{}}
\toprule
\multicolumn{1}{c}{Hint} & \begin{tabular}[c]{@{}c@{}}Percentage of Completely\\ Incorrect Data (\%)\end{tabular} & \begin{tabular}[c]{@{}c@{}}Average Score for \\ Four Attempts\end{tabular} & Standard Deviation \\ \midrule
no hint & 25.56 & 1.62 & 1.18 \\
level 1: conceptual & 13.44 & 2.53 & 1.43 \\
level 2: strategic & 12.25 & 2.66 & 1.41 \\
level 3: tooling & 10.08 & 2.70 & 1.39 \\
level 4: procedural & 8.96 & 2.87 & 1.35 \\
level 5: solution & 0.20 & 3.96 & 0.28 \\ \bottomrule
\end{tabular}%
\end{table}

\subsubsection{Effect of Partial Solutions During RL Phase}
To evaluate the model's performance under different hint conditions, the success rate of solving problems across four rollout attempts serves as the primary metric. Approximately 8,000 RL data samples are selected for testing, with the following conditions:
\begin{itemize}
    \item The model samples the same problem independently four times under each hint condition.
    \item The score is determined by the number of correct answers (ranging from 0 to 4).
\end{itemize}
As shown in Table \ref{tab:partial_solution}, without any hints, approximately 25.56\% of the samples fail to provide a correct solution, significantly reducing the efficiency of the RL process. As the hints approach a complete solution (level 5), the error rate decreases, and the average score for the four attempts increases, indicating more stable and accurate responses. Additionally, a comparison between performance in the RL phase with and without partial solutions in Table \ref{tab:training_comparison} shows improvements across various benchmarks, including an increase in the average score from 79.41 to 80.13 on OpenCompass, and a 1.3-point improvement on MathVista, further validating the effect of partial solutions.


\subsubsection{Impact of Rejection Sampling on SFT and RL Performance}
\begin{figure*}[t]
\centering
\includegraphics[width= 1\linewidth]{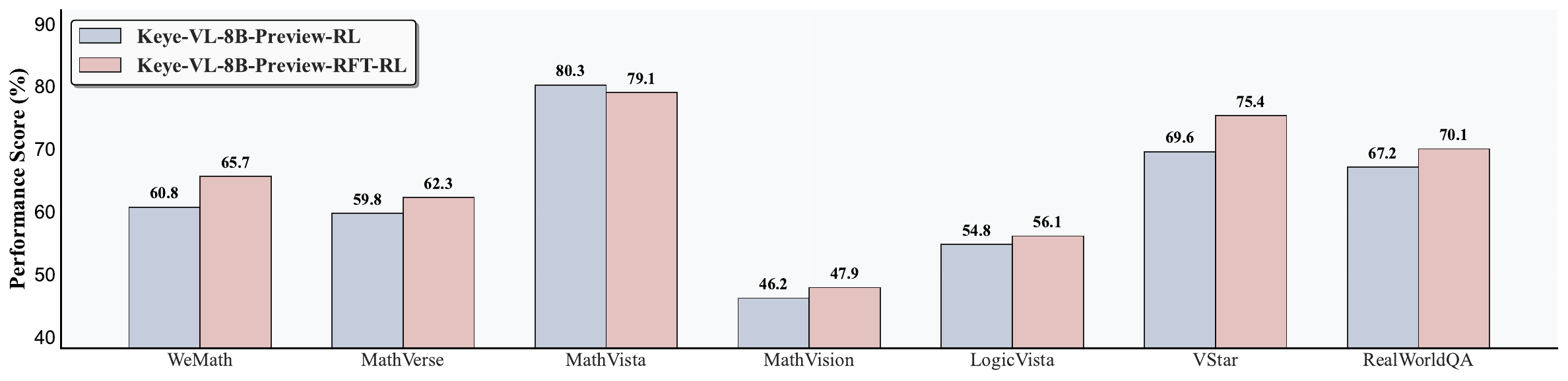}
\caption{\textbf{Benefits of rejection sampling in the RL Phase.} Starting from Keye-VL-8B-Preview, we compare the performance of direct RL and RFT-RL strategies.}
\label{fig:rft_rl}
\end{figure*}

In our RL iteration process, we employ rejection sampling twice. To validate the effectiveness of this approach, we conduct experiments starting with Keye-VL-8B-Preview, training it with the same RL dataset. In contrast, Keye-VL-8B-Preview-RFT-RL undergoes one round of iteration, followed by a second RL training phase. As shown in Figure~\ref{fig:rft_rl}, this iterative strategy significantly boosts RL performance, increasing the average mathematical benchmark score from 60.37 to 62.24, with similar improvements observed across general reasoning benchmarks. In Table~\ref{tab:training_comparison}, we compare the impact of various strategies, including Long CoT Cold Start, rejection sampling of SFT data using an RL model, and the subsequent selection of the best samples using a reward model for further SFT training (RFT-SFT). As a result, OpenCompass's average score rises from 75.32 to 76.33, with consistent performance improvements across other benchmarks. Based on these findings, we adopt the SFT-RL-(RFT-SFT)-(RFT-RL) iterative model to further enhance performance.

\section{Conclusion and Discussion}
In this work, we presented Keye-VL-1.5, an advanced multimodal model that significantly enhances video understanding and vision-language tasks. By employing a novel Slow-Fast video encoding strategy, we efficiently balance temporal coverage and spatial resolution. The model’s progressive pre-training, with an extended context length, enables it to handle longer videos and complex visual content, while post-training methods focused on reasoning and human preference alignment improve instruction-following and reasoning abilities. Our evaluation demonstrates that Keye-VL-1.5 advances video understanding capabilities while maintaining strong performance on general vision-language tasks.

\bibliography{colm2024_conference}
\bibliographystyle{unsrtnat}

\newpage
\appendix

\newpage

\section{Case Study}




\label{video_cases}

\subsubsubsection{Video Case I: Grounding}
\label{case_video_ground}
\renewcommand{\arraystretch}{1.5}
\begin{figure}[!h]
  \centering
  \begin{tabular}{m{16cm}}
  \toprule
  \begin{center}
  Input Video
  \end{center}
  \begin{center}
  \includegraphics[width=0.8\linewidth]{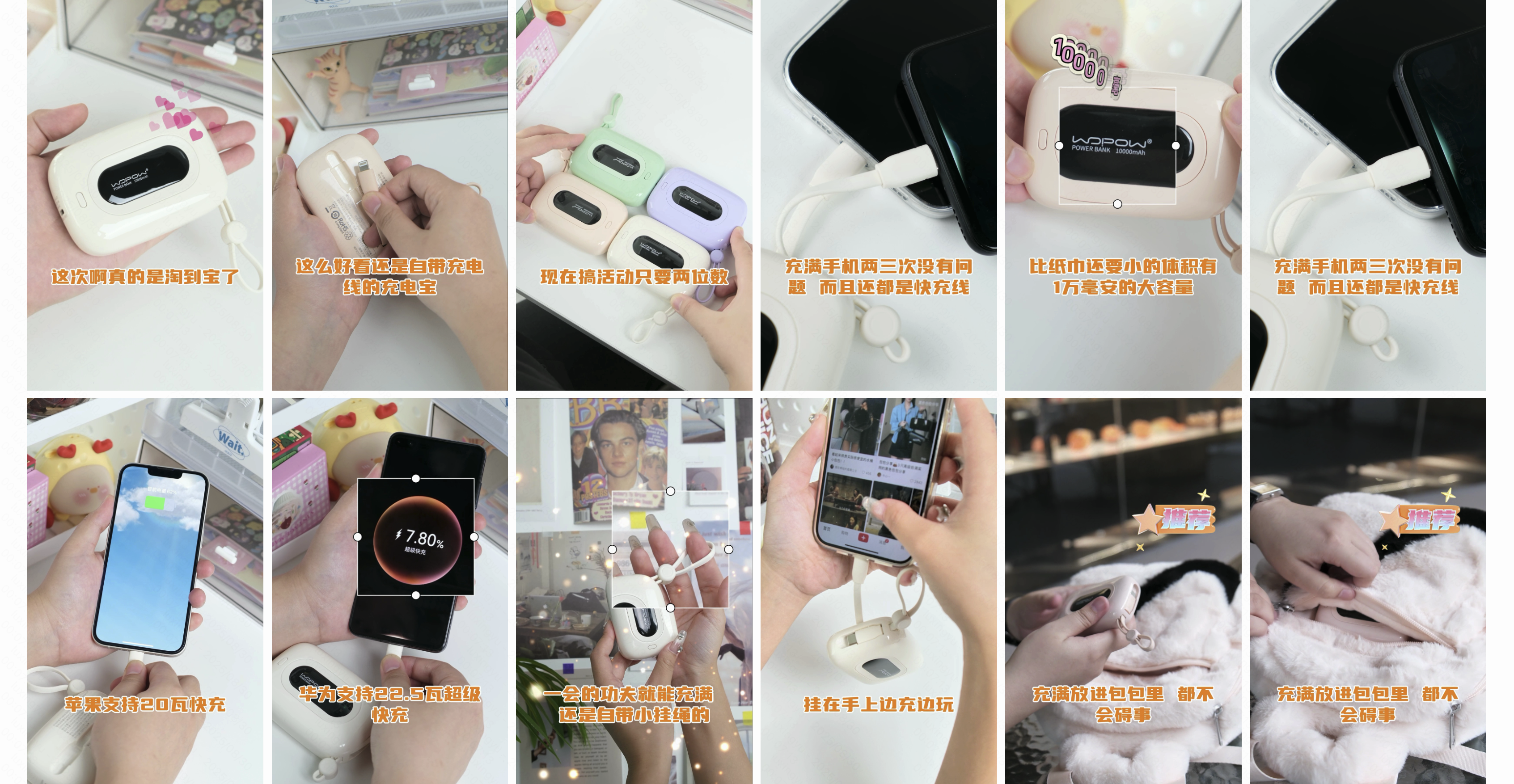}
  \end{center}
  \\
{\footnotesize \textbf{Prompts:} What is the time period during which the handbag appears in this video?\newline\newline
\textbf{Model Responses:} {The time period during which the bag appears in the video is from 22.3 seconds to 23.8 seconds, i.e., \boxed{22.3s - 23.8s}.}\newline
}
  \\
  \bottomrule
  \end{tabular}
\caption{In this 26-second video, the creator provides a detailed introduction to the performance, advantages, and features of the power bank. The handbag only appears for about two seconds in the last of the video. Keye-VL-1.5 accurately pinpoints this time range, with a precision of 0.1 seconds.}
\label{fig:video_ground}
\end{figure}
\renewcommand{\arraystretch}{1}

\clearpage
\subsubsubsection{Video Case II: Content Understanding}
\label{case_video_understand}
\renewcommand{\arraystretch}{1.5}
\begin{figure}[!h]
  \centering
  \begin{tabular}{m{16cm}}
  \toprule
  \begin{center}
  Input Video
  \end{center}
  \begin{center}
  \includegraphics[width=0.8\linewidth]{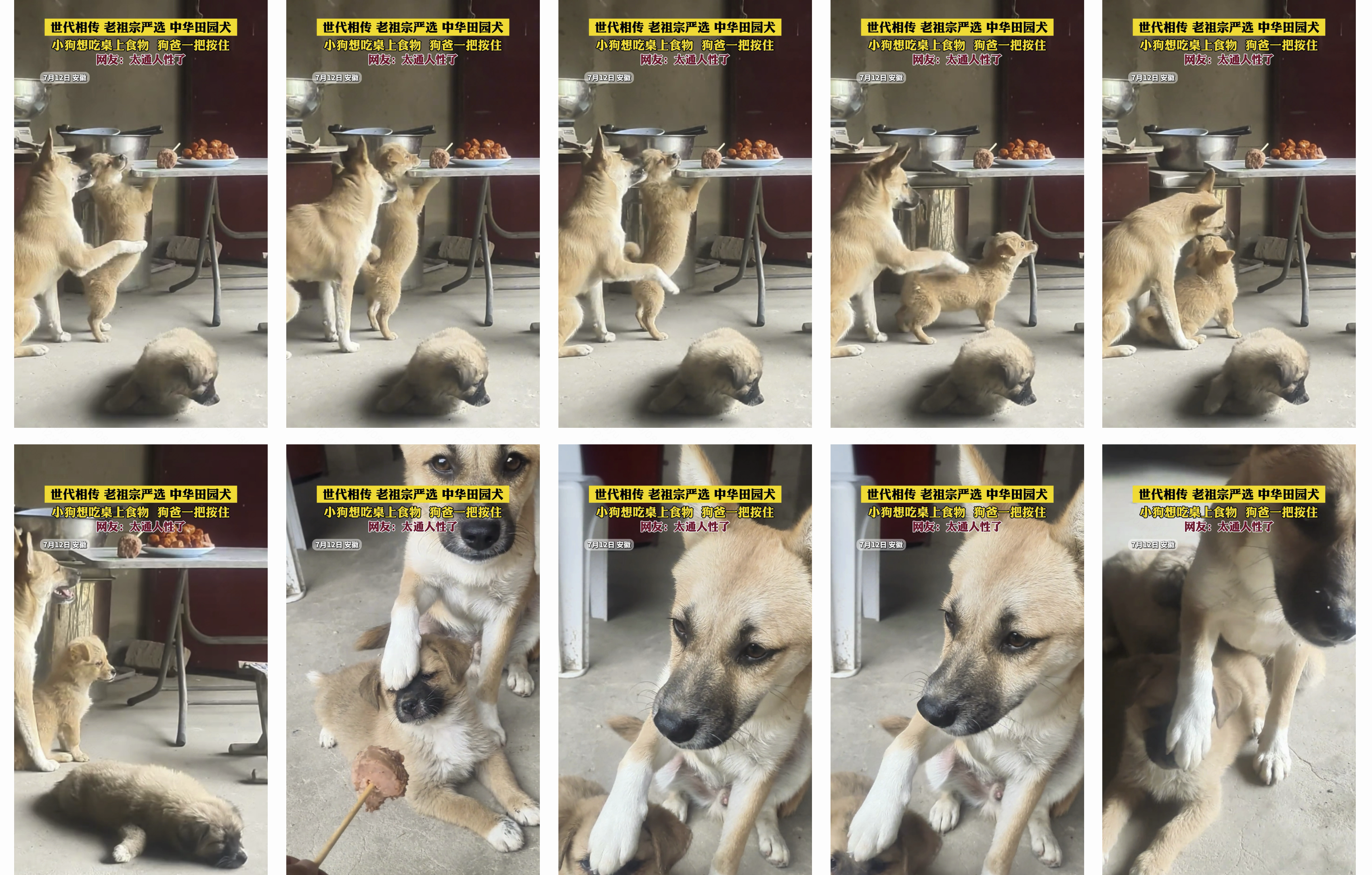}
  \end{center}
  \\
{\footnotesize \textbf{Prompts:} What is the purpose of the big dog biting the little dog's ear in this video?\newline\newline
\textbf{Model Responses:} {<think> In the video, the little dog tries to reach the food on the table, and the big dog (the "dad") lightly bites the little dog's ear as a way of \textbf{correcting the little dog's improper behavior}. This gentle bite is a common "educational" method in dog packs, where adult dogs use it to tell the puppies, "This is wrong," and that rules must be followed. After receiving this signal, the little dog gradually sits down from the standing position and eventually lies down to comply, indicating that the big dog's action successfully conveyed the warning of "don't steal food from the table."</think> \newline
<answer>The big dog bites the little dog's ear to correct its improper behavior of trying to steal food from the table, using a gentle bite to deliver a warning and make the little dog comply.</answer>
}\newline
}
  \\
  \bottomrule
  \end{tabular}
\caption{In this video, the caption shows that the dog dad prevents the little dog from eating the food on the dining table by holding the little dog down. The question asks about the reason behind the big dog biting the little dog's ear. Keye-VL-1.5 accurately correlates the video content and the caption, determining that the purpose of the big dog biting the little dog's ear is the same as holding the little dog down – both are meant to teach the little dog not to eat the food on the table.}
\label{fig:video_understand}
\end{figure}
\renewcommand{\arraystretch}{1}

\clearpage
\subsubsubsection{Video Case III: Content Description}
\label{case_video_desc}
\renewcommand{\arraystretch}{1.5}
\begin{figure}[!h]
  \centering
  \begin{tabular}{m{16cm}}
  \toprule
  \begin{center}
  Input Video
  \end{center}
  \begin{center}
  \includegraphics[width=0.8\linewidth]{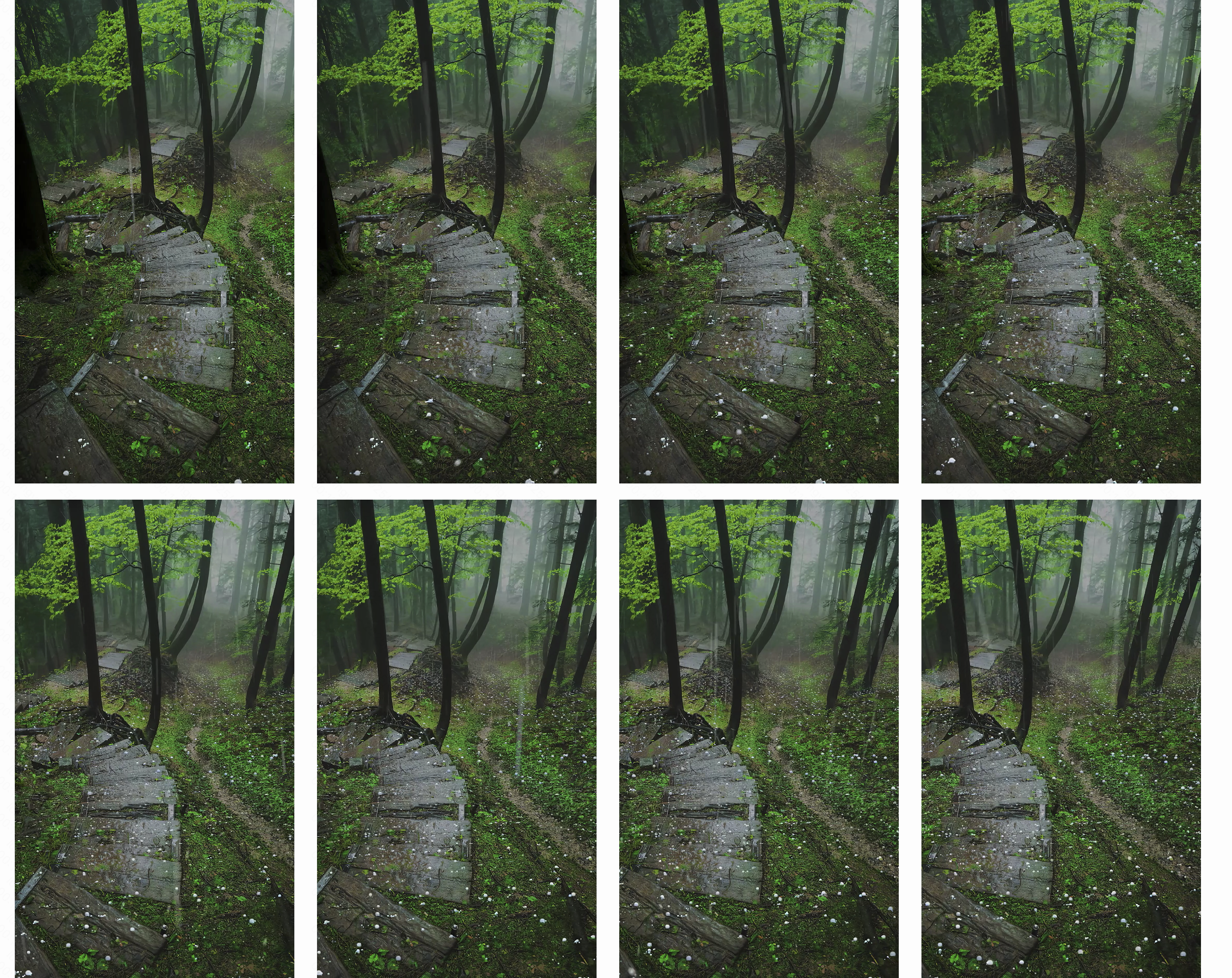}
  \end{center}
  \\
{\footnotesize \textbf{Prompts:} Please describe the scene in the video in detail.\newline\newline
\textbf{Model Responses:} {The main scene of the video is a damp corner surrounded by dense forest. In the center of the frame is an old, broken stone staircase, with cracks in the stones and the surface covered in green moss and weeds, giving it a weathered and worn appearance. Surrounding it are tall, dark trees with branches hanging with fresh green leaves. Soft, diffused light filters through the leaves, adding a layer of misty ambiance to the scene. The ground is scattered with small white flowers or fungi, interwoven with the green moss, forming a vibrant, living ground cover. In the background, the forest stretches into the distance, with trees becoming gradually blurred, adding depth and mystery to the scene. The air seems to be filled with fine mist, making the ground and the stone steps appear damp, with occasional raindrops falling, further enhancing the humid atmosphere. The overall color tone is dominated by greens and browns, presenting a serene, natural, and slightly desolate image of a primal forest.}\newline
}
  \\
  \bottomrule
  \end{tabular}
\caption{This video describes a very beautiful yet rare phenomenon: hail falling in the forest. Without prior knowledge, even human viewers could easily mistake the falling hail for petals or raindrops. Although Keye-VL-1.5 fails to identify the falling objects as hail, it mentions precipitation, and the rest of the content is correctly identified.}
\label{fig:video_desc}
\end{figure}
\renewcommand{\arraystretch}{1}

\clearpage

\clearpage


\section{Authors (Alphabetical order)}
\textbf{Core Contributors:} Biao Yang, Bin Wen, Boyang Ding, Changyi Liu, Chenglong Chu, Chengru Song, Chongling Rao, Chuan Yi, Da Li, Dunju Zang, Fan Yang, Guorui Zhou, Guowang Zhang, Han Shen, Hao Peng, Haojie Ding, Hao Wang, Haonan Fan, Hengrui Ju, Jiaming Huang, Jiangxia Cao, Jiankang Chen, Jingyun Hua, Kaibing Chen, Kaiyu Jiang, Kaiyu Tang, Kun Gai, Muhao Wei, Qiang Wang, Ruitao Wang, Sen Na, Shengnan Zhang, Siyang Mao, Sui Huang, Tianke Zhang, Tingting Gao, Wei Chen, Wei Yuan, Xiangyu Wu, Xiao Hu, Xingyu Lu, Yi-Fan Zhang, Yiping Yang, Yulong Chen, Zeyi Lu, Zhenhua Wu, Zhixin Ling, Zhuoran Yang, Ziming Li.

\textbf{Contributors:} Di Xu, Haixuan Gao, Hang Li, Jing Wang, Lejian Ren, Qigen Hu, Qianqian Wang, Shiyao Wang, Xinchen Luo, Yan Li, Yuhang Hu, Zixing Zhang.

\end{document}